\pdfoutput=1
\documentclass[11pt]{article}

\usepackage[final]{acl}

\usepackage{times}
\usepackage{latexsym}
\usepackage{booktabs}
\usepackage{multirow}
\usepackage{multicol}

\usepackage[T1]{fontenc}
\usepackage[utf8]{inputenc}

\usepackage{microtype}

\usepackage{inconsolata}
\usepackage{amsfonts,amssymb}

\usepackage{graphicx}
\usepackage{multirow}
\usepackage{pifont}
\usepackage{xcolor}
\usepackage{booktabs}
\usepackage{hyperref}
\usepackage{makecell}
\usepackage{amsmath}
\usepackage{amsfonts}
\usepackage{enumitem}
\usepackage{multicol}
\usepackage[ruled,vlined]{algorithm2e}
\usepackage{algpseudocode}
\usepackage{soul}
\usepackage[most]{tcolorbox}
\tcbuselibrary{skins}
\tcbuselibrary{breakable}

\definecolor{mintleaf}{RGB}{0, 184, 148}
\definecolor{dm-blue-500}{RGB}{0, 69, 177}
\definecolor{dm-purple-500}{RGB}{105,50,230}
\definecolor{mysilver}{RGB}{128,129,128}
\definecolor{my_green}{RGB}{0, 176, 80}
\definecolor{my_yellow}{RGB}{255,165,0}
\definecolor{my_red}{RGB}{255, 0, 0}
\definecolor{my_purple}{RGB}{126, 100, 158}
\definecolor{my_blue}{RGB}{49, 133, 155}
\definecolor{case_purple}{RGB}{160, 43, 147}
\definecolor{case_blue}{RGB}{15, 158, 213}

\newcommand{\ourmethod}{IPR} 
\title{Watch Every Step! LLM Agent Learning via  \\ Iterative Step-Level Process Refinement}
\author{%
  Weimin Xiong$^1$,
  Yifan Song$^1$,
  Xiutian Zhao$^2$,
  Wenhao Wu$^1$,
  Xun Wang$^1$\\
  \textbf{Ke Wang}$^2$,
  \textbf{Cheng Li}$^2$,
  \textbf{Wei Peng}$^2$,
  \textbf{Sujian Li}$^1$\thanks{Corresponding Authors.}\\
  $^1$State Key Laboratory of Multimedia Information Processing\\ School of Computer Science, Peking University\quad\\
  $^2$Huawei Technologies\quad\\
  \texttt{\{wmxiong, lisujian\}@pku.edu.cn} \\
  \vspace{-3mm}\\
}

\begin{document}
\maketitle
\begin{abstract}
% Large language model~(LLMs) agents have demonstrated outstanding performance in various complex interaction tasks. While recent efforts have used trajectory-based tuning to enhance agent performance, they focus solely on outcome rewards. In our work, we introduce the Iterative Process Refinement (IPR) framework, which incorporates process-level refinement into agent training, enabling learning contrastive action pairs. In each iteration, the agent explores along the expert trajectory to generate new actions, and we use sampling to estimate step rewards, thereby constructing contrastive data. Then during the training phase, we apply mixture trajectory optimization to optimize the agent, providing process supervision. Our experimental results on three agent benchmarks consistently show that our method outperforms other baselines. Further analysis confirms that our method increases the agent's average step reward and is applicable to various base models.

Large language model agents have exhibited exceptional performance across a range of complex interactive tasks. Recent approaches have utilized tuning with expert trajectories to enhance agent performance, yet they primarily concentrate on outcome rewards, which may lead to errors or suboptimal actions due to the absence of process supervision signals. In this paper, we introduce the \textbf{I}terative step-level \textbf{P}rocess \textbf{R}efinement \textbf{(\ourmethod{})} framework, which provides detailed step-by-step  guidance to enhance agent training. 
Specifically, we adopt the Monte Carlo method to estimate step-level rewards. During each iteration, the agent explores along the expert trajectory and generates new actions. These actions are then evaluated against the corresponding step of expert trajectory using step-level rewards. Such comparison helps identify discrepancies, yielding contrastive action pairs that serve as training data for the agent. 
Our experiments on three complex agent tasks demonstrate that our framework outperforms a variety of strong baselines. Moreover, our analytical findings highlight the effectiveness of \ourmethod{} in augmenting action efficiency and its applicability to diverse models\footnote{Code \& Data: \href{https://github.com/WeiminXiong/IPR}{https://github.com/WeiminXiong/IPR}.}.

\end{abstract}

% 两个图要更新

\section{Introduction}
\begin{figure}[ht]
    \centering
    \includegraphics[width=\linewidth]{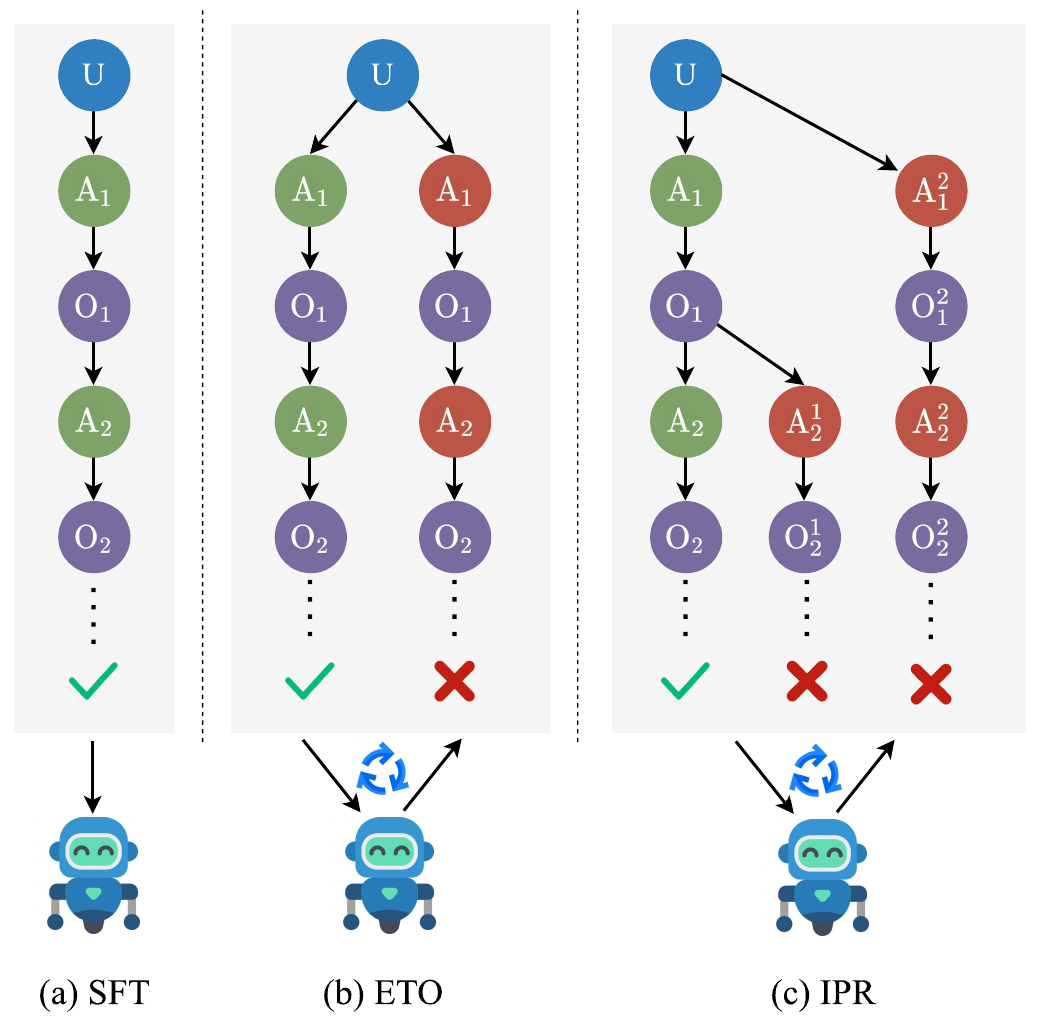}
    \caption{Comparison of three different agent training paradigms. Green and red circles represent correct and incorrect actions, while check and cross marks indicate the final outcome. Compared to the other methods, \ourmethod{} can provide step-level process supervision.}
    \label{fig:overall}
\end{figure}
% 背景介绍
% With the rapid development of artificial intelligence, large language models (LLMs) like GPT-3.5~\citep{ouyang2022training}, GPT-4~\citep{achiam2023gpt}, LLaMA~\citep{touvron2023llama} mark a sigficant shift in language understanding, reasoning and generation. The advancements in large language models have opened doors for agents to excel in handling complex interactive tasks. 
% From online shopping ~\citep{yao2022webshop} to SQL database querying ~\citep{yang2024intercode} and textual embodied ~\citep{shridhar2020alfworld}, LLMs empower agents to effortlessly interact with the environment, automating tasks with impressive proficiency.

The advancements in large language models (LLMs), such as GPT-3.5~\citep{ouyang2022training}, GPT-4~\citep{achiam2023gpt}, LLaMA~\citep{touvron2023llama} have paved ways for LLM-based agents to excel in handling complex interactive tasks, including
online shopping~\citep{yao2022webshop} and embodied housework~\citep{shridhar2020alfworld}. 
% The task-solving process, where an agent interacts with the environments step by step, is the main concern and critical to the final performance of the agent.
To accomplish these tasks, LLM agents explore the environment step by step, achieving sub-goals along action trajectories~\citep{ma2024agentboard}. The efficacy of this task-solving process is pivotal to agent's overall performance.

Initial efforts in the task-solving process for agents involve generating trajectories by  directly leveraging the planning ability of LLMs, such as ReAct~\citep{yao2022react} and Reflexion~\citep{shinn2024reflexion}.  
To further enhance LLM agent abilities, several studies focus on trajectory tuning~\citep{chen2023fireact,yin2023lumos,zeng2023agenttuning}. 
\citet{chen2023fireact} and \citet{yin2023lumos} construct agent trajectory data from teacher agents (e.g., GPT-4) and fine-tune open-source LLMs for specific agent abilities, such as reasoning. Conversely, \citet{zeng2023agenttuning} employ a multi-task supervised fine-tuning (SFT) approach, which does not significantly improve generalized agent capabilities.
Observing that the SFT-based works predominantly rely on expert success trajectories (Figure \ref{fig:overall}(a)), \citet{song2024trial} utilize failure trajectories and propose the exploration-based trajectory optimization (ETO) method to learn the task-solving process (Figure \ref{fig:overall}(b)).
Although these methods present a promising avenue for enhancing agent capabilities, they treat an entire trajectory as a single entity during training and  
% value the outcome of a trajectory more than process, 
prioritize the final reward of a trajectory over the process,
% causing the neglect of the valuable process information.
thus overlooking the potentially exploitable information throughout interaction process.

Regarding agent trajectories, it is well-known that alongside those with correct outcomes, there are trial-and-error paths with detours and erroneous ones that achieve accidental success. Step-level process supervision can offer granular guidance at each step hence is beneficial for task resolution ~\citep{lightman2023let}. 
% To the best of our knowledge, no work has applied step-level supervision on improving LLM agents, as there are two challenges below.
Nevertheless, the application of step-level optimization to LLM agents encounters two practical challenges.
%Step-level Process supervision can provide agents with more granular feedback~\citep{lightman2023let}, offering detailed guidance at each step. Intuitively, if an agent can learn precisely where it made an error in the process, it can choose the most accurate path to complete the task (Fig~\ref{fig:overall}).
%However, there are two main challenges in implementing this.
Firstly, the majority of existing LLM agent environments~\citep{yao2022webshop, shridhar2020alfworld, yang2024intercode} provide only final outcome feedback. Even in cases where environments offer sub-goal level feedback ~\citep{ma2024agentboard}, the information is often too sparse. 
% Second, there is the challenge of collecting step-level training data for agents and utilizing it effectively to enhance the training.
Secondly, the question of how to effectively utilize step rewards to enhance agent training, particularly for tasks with long trajectories and complex action spaces, remains unexplored.

In this paper, we address these challenges by introducing the \textbf{I}terative step-level \textbf{P}rocess \textbf{R}efinement \textbf{(\ourmethod{})} framework (\S~\ref{sec:method}) , which encompasses two principal mechanisms: Step-level Reward Acquisition (\S~\ref{sec:reward estimation}) and Iterative Agent Optimization (\S~\ref{sec:training}). More specifically, to construct the step reward within the agent environment, we 
% apply Monte Carlo search 
employ Monte Carlo~(MC) method to estimate rewards via sampling. 
The Iterative Agent Optimization component aims to refine the agent's actions through a cyclical process. During each cycle, the agent navigates the expert trajectory and generate new actions. These actions are then compared with the corresponding step of the expert trajectory using step-level rewards to pinpoint errors, resulting in contrastive step pairs.
Subsequently, we train the agent using an arrangement of outcome-level direct preference optimization (DPO), step-level DPO, and SFT losses, thereby enhancing the agent's action capabilities at each step~(Figure \ref{fig:overall}(c)).

% 摆一些实验结果
We assess our \ourmethod{} framework on three representative benchmarks: online shopping environment WebShop~\citep{yao2022webshop}, interactive SQL environment InterCodeSQL~\citep{yang2024intercode} and textual embodied environment ALFWorld~\citep{shridhar2020alfworld}. The experimental results, detailed in \S~\ref{sec:results}, reveal that our method surpasses the current leading method by margins of 5.8\%, 7.2\% and 3.2\% on WebShop, InterCodeSQL, and ALFWorld, respectively.
% Specifically, in the WebShop scenario, our method outperforms the state-of-art  by 4.2\%, while in ALFWorld, it achieves a 3.2\% improvement. 
% Further analyses show that \ourmethod{} is applicable to different LLMs, accurately estimating step rewards and increasing the average reward for each action taken by the agent. 
% Moreover, further analytical results show that: 1) our \ourmethod{} can be applied to various base LLMs, 2) our estimation of the step reward is relatively accurate, 3) our \ourmethod{} increases the reward per step for the agent, improving the efficiency of task completion.
% Moreover, additional analyses (\S~\ref{sec:analysis}) indicate that: (1) our \ourmethod{} enhances the reward per step for the agent, thereby increasing the efficiency of task completion; and (2) constructing a step reward model automatically is a viable approach to reduce the training costs associated with the MC method.
Moreover, we present a comprehensive analysis to substantiate the efficacy of our method from various perspectives.

% Contridution
In summary, our contributions are as follows:

\begin{itemize}[leftmargin=*, nolistsep]
\setlength{\itemsep}{1mm}
% \item  We introduce the \ourmethod{} framework, which enables agents to autonomously learn from step-wise trajectories they explore and obtain fine-grained reward information for each step in the trajectory.
\item  We introduce the \ourmethod{} framework, marking the first integration of step-level process supervision into LLM agent training. This innovation enables fine-grained adjustments of the agent's task completion.
% \item \textbf{Performance.} Our experiments across three complex interactive tasks reveal that \ourmethod{} outperforms established leading baselines.
\item Our experiments across three complex interactive agent tasks reveal that \ourmethod{} outperforms established leading baselines.
% \item \textbf{Analysis.} We present a comprehensive analysis to substantiate the efficacy of our method from various perspectives and explore the feasibility of automatically constructing step reward models for agent environments.
\item Additional analyses indicate that: (1) our \ourmethod{} enhances the reward per step for the agent, thereby increasing the efficiency of task completion; and 
(2) constructing a step reward model automatically is a viable approach to reduce the training costs associated with the MC method.
\end{itemize}

\begin{figure*}[!htbp]
    \centering
    \includegraphics[width=\textwidth]{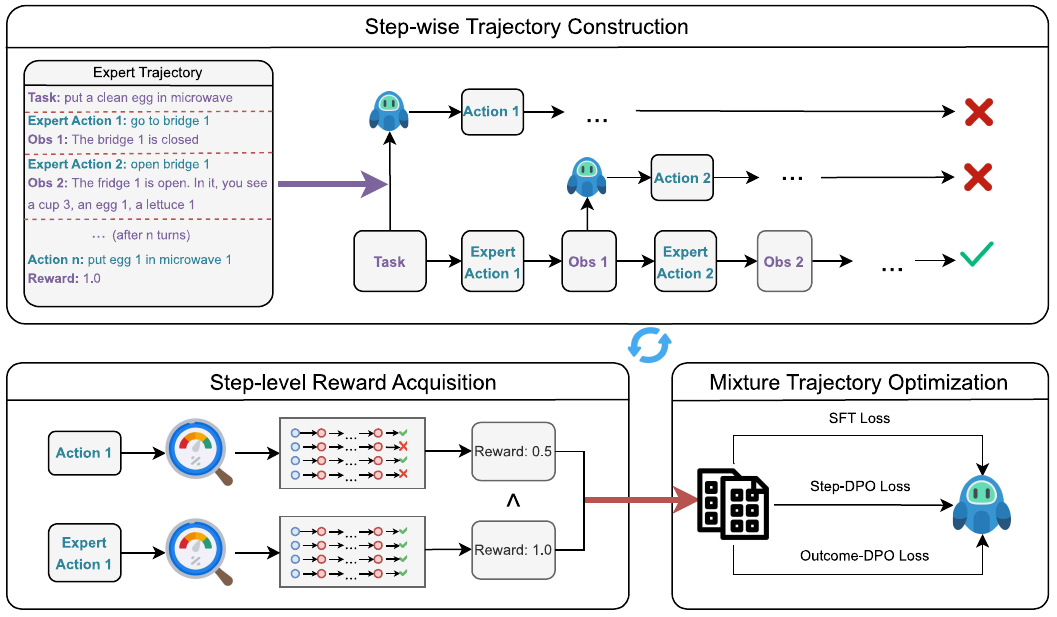}
    \caption{The overall architecture of \ourmethod{} in a single iteration. The agent trained after SFT first explores new actions along the expert trajectory. Then we use the scorer to reward each step and construct contrastive action data. Finally we optimize the agent with a mixed loss.}
    \label{fig:structure}
\end{figure*}

\section{Task Formulation}
% The agent task with environment feedback can be formalized as a partially observable Markov decision process (POMDP) ($\mathcal U, \mathcal S, \mathcal A, \mathcal O, \mathcal T, \mathcal R$) with instruction space $\mathcal U$, state space $\mathcal S$, action space $\mathcal A$, observation space $\mathcal O$, transition function $\mathcal T: \mathcal S \times \mathcal A \rightarrow \mathcal S$, and reward function $\mathcal R: \mathcal S \times \mathcal A \rightarrow [0, 1]$. Note in our LLM-based agent scenario, $\mathcal U, \mathcal A, \mathcal O$ are subsets of natural language space.

The primary scope of this study is the task-solving of LLM agents interacting with the environment and receiving feedback. 
Following~\citet{song2024trial}, we formulate the task as a partially observable Markov decision process (POMDP) defined by the elements ($\mathcal U, \mathcal S, \mathcal A, \mathcal O, \mathcal T, \mathcal R$). Here, $\mathcal U$ denotes the instruction space, $\mathcal S$ the state space, $\mathcal A$ the action space, $\mathcal O$ the observation space, $\mathcal T$ the transition function ($\mathcal T: \mathcal S \times \mathcal A \rightarrow \mathcal S$), and $\mathcal R$ the reward function ($\mathcal R: \mathcal S \times \mathcal A \rightarrow [0, 1]$). In the context of our LLM-based agent, $\mathcal U, \mathcal A, \mathcal O$ are subsets of natural language space.

% In our task, there are two different agents parameterized as $\pi_\theta$ and $\pi_e$.
% , which can generate the action $a_t \in \mathcal A$ at step $t$. 
% The difference between them is the different roles they play in the framework. The base agent that we aim to optimize, $\pi_\theta$, continuously interacts with the environment and learns from the trajectories. While the explorer, $\pi_e$, has fixed parameters and conducts step-level exploration to identify the error step. 
% The interaction loop is demonstrated as follows: 
% $$e = (u, a_1, o_1, ..., o_{n-1}, a_n) \sim \pi(e|u), $$
% $$\pi(e|u) = \prod \limits_{i=0}^n \pi(a_i|u, a_1, o_1, ..., o_{i-1}), $$ where n is the trajectory length. Specifically, the agent generates the next action $a_{i+1}$ based on the instruction $u$ and historical actions $(a_1, a_2,..., a_{i-1})$ and environmental feedbacks as observations $(o_1, o_2, ...o_{i-1})$. Finally, the final reward $r(u, e) \in [0, 1]$ is computed.
% , which will be used to compute the step reward.
At time step $t$, the LLM agent $\pi_\theta$ receives the observation $o_{t-1} \in \mathcal O$ from the environment and takes an action $a_t \in \mathcal A$ following the policy $\pi_\theta(\cdot |e_{t-1})$, where $e_{t-1} = (u, a_1, o_1, ..., a_{t-1}, o_{t-1})$ represents the historical trajectory. The action leads to a change in the state space $s_t \in \mathcal S$, and receives execution feedback as observation $o_t \in \mathcal O$. The interaction loop continues until the task is completed or the maximum steps are reached.
The final trajectory is $e_n = (u, a_1, o_1, ..., a_n, o_n)$, where $n$ denotes the trajectory length, and the outcome reward is $r_o(u, e_n) \in [0, 1]$.
For the convenience of subsequent content, we define $e_{t:n} = (a_t, o_t, ..., a_n, o_n)$ to represent the trajectory after time step $t$.

\section{Method}\label{sec:method}
% In this section, we will introduce the details of our proposed framework, which is composed of three parts.
The overall architecture of our method is depicted in Figure~\ref{fig:structure}.
Initially, we empower the language model with fundamental agent capabilities via supervised learning~(\S~\ref{sec:sft}). 
% Then, we use different prompts to get the agent to generate the subsequent trajectories. 
% Then, we construct contrastive action pairs based on the reward for each step.
Subsequently, we develop the MC method to estimate the step-wise rewards within the agent's environment~(\S~\ref{sec:reward estimation}). 
% And, we use a monte-carlo based sampling method to get each step reward. 
% Finally, we construct preference pairs based on the reward for each step and use the preference pairs to further improve the agent's abality on the corresponding task.
In the final stage, we enhance the agent's performance through iterative optimization~(\S~\ref{sec:training}): by constructing contrastive action pairs and executing mixture trajectory optimization. 
% Lastly, we optimize the performance of the agent iteratively~(\S~\ref{sec:training}) via  constructing contrastive action pairs and performing mixture trajectory optimization.

\subsection{Supervised Fine-tuning}
\label{sec:sft}

To develop an agent with basic task capabilities, we perform supervised fine-tuning (SFT) on an expert trajectory dataset in ReAct-Style~\citep{yao2022react}. We denote this expert trajectory as $\mathcal D = \Big\{(u, e)^{(i)}\Big\}_{i=1}^{|\mathcal D|}$, where $|\mathcal D|$ is the number of trajectories. The loss can be computed as:
\begin{equation}
    \mathcal L_{\mathrm SFT}(\theta) = -\mathbb E_{e \sim \mathcal D}[\log \pi_\theta(e|u)].
\end{equation}
Since $\pi_{\theta}(e|u) = \prod_{t=1}^n\pi_{\theta}(a_t|u, ..., o_{t-1}) = \prod_{t=1}^n\pi_{\theta}(a_t|e_{t-1})$ in practice. The loss function can further be expressed as:
\begin{equation}
    \mathcal L_{\mathrm SFT}(\theta) = -\mathbb E_{e \sim \mathcal D}\bigg[\sum_{t=1}^n \log \pi_\theta(a_t|e_{t-1})\bigg].
\end{equation}

% 3.1 SFT
% 3.2 Step-wise reward acquiring
% why to use step-wise reward/information.
% get step-wise reward function
% 3.3 Coarse-to-fine/Mixture/Multi-Level/Hybrid trajectory optimization
% RL instable, so we use DPO
% \pararaph{} DPO step-wise contrastive data construction
% \pararaph{} DPO optimization

\subsection{Step-level Reward Acquisition}
\label{sec:reward estimation}
Step-level process reward provide precise feedback by pinpointing the exact location of potential errors, offering a valuable signal for agent learning. However, most agent environments are limited to outputting only final outcome reward. Prior studies~\citep{uesato2022solving, lightman2023let} rely on human annotators for step supervision annotations, rendering the acquisition of step rewards a labor-intensive process. 
% These annotation can be an arduous process and require advanced annotator skills which is quite costly.
% To tackle this problem, we draw inspiration from the Monte Carlo Tree Search~(MCTS)~\citep{kocsis2006bandit, coulom2006efficient} algorithm to estimate the reward for each action.
% To address this, we use an exploration-based method to estimate the reward for the action $a_t$ at step $t$.
To circumvent this, we adopt an exploration-based method to estimate the reward for action $a_t$ at step $t$.

% For a trajectory $e = (u, a_1, o_1, ..., a_n, o_n)$, to estimate the step reward, 
% Intuitively, if a particular action is more correct, it is more likely to contribute to achieving a higher reward outcome.
It is intuitive that a more accurate action would contribute to a higher reward.
% Therefore, to estimate the step reward at step $t$ for a trajectory $e = (u, a_1, o_1, ..., a_n, o_n)$, 
% Therefore, we represent the reward of the action $a_t$ as the expected outcome reward by exploring from the current step onward.
Therefore, we define the step reward $r_s(s_t, a_t)$ as the anticipated outcome reward from subsequent exploration starting at step $t$, with $s_t$ being the current state of the environment. 
% we define the reward of the action $a_t$ as the expected outcome reward obtained by starting from state $s_t$, 
% We denote the reward function at step $t$ as $r_s(s_t, a_t)$.
% where $s_t$ is the state of the environment at step $t$. 
% we define the reward function at step $t$ as $r_s(s_t, a_t)$, 
% where $s_t$ is the state of the environment at step $t$. 
% When the task is completed, the environment can provide an outcome reward $r_o(u, e)$. 
% Specifically, we use an agent to explore from the current step $t$, referred to as the explore agent $\pi_e$, with its parameters fixed. 
% Using $e_{t-1}$ as  history, $\pi_e$ can interact with the environment to sample a new trajectory $e_m$
A dedicated scorer $\pi_s$ with fixed parameters is employed to generate new subsequent trajectory $e_{t:m}$ from step $t$, based on the historical trajectory $e_{t-1}$. 
% \begin{equation}
%     \hat{e}_m = (u, ..., \hat{a}_t, \hat{o}_t, ..., \hat{o}_m, \hat{a}_m) \sim \pi_e(\hat{e}_m|e_{t-1})
% \end{equation}
% $\hat{e}_m = (u, a_1, o_1, ..., o_{t-1}, \hat{a}_t, \hat{o}_t, ..., \hat{o}_m, \hat{a}_m)$ 
% with the probability represented as $\pi_e(e_m|e_{t-1})$. 
The probability of generating $e_{t:m}$ is given by $\pi_s(e_{t:m}|e_{t-1})$, and the environment assigns an outcome reward $r_o(u, e_m)$ for the trajectory.
% For this trajectory, the environment will return an outcome reward $r_o(u, e_m)$. Therefore, the step reward $r_s(s_t, a_t)$ can be calculated as:
The step reward can be calculated as:
\begin{equation}
    r_s(s_t, a_t) = \mathbb E_{e_m \sim \pi_s(e_{t:m}|e_{t-1})} [r_o(u, e_m)]
\end{equation}
% However, it is difficult to provide a closed-form solution for this expectation. Therefore, we use sampling to estimate the expectation.
% We use the explore agent $\pi_e$ to sample $N$ trajectories from step $t$ , resulting in the set of trajectories
Given the complexity of directly calculating this expectation value, we employ Monte Carlo sampling method for estimation.
By sampling $N$ trajectories from step $t$ with $\pi_s$, we generate a set of trajectories:
% $\{\hat{e}_{m_i}\}_{i=1}^N$
% \begin{equation}
%     \{e_m^1, ...,e_m^N\} = MC^{\pi_s}(e_{t-1}; N),
% \end{equation}
\begin{equation}
    \{e^{(i)} | i=1,...,N \} = MC^{\pi_s}(e_{t-1}; N),
\end{equation}
% In this work, we directly use the model trained through SFT, which has the ability to perform the corresponding task, as our explore agent $\pi_e$. Afterward, we can estimate the step reward as:
The step reward is then calculated as:
% \begin{equation}
%     r_s(s_t, a_t) = \left\{
%     \begin{aligned}
        
%     \end{aligned}\frac{1}{N} \sum_{i=1}^N r_o(u, \hat{e}_{m_i}).
% \end{equation}
% \begin{equation}
% r_s(s_t, a_t) =
%     \begin{cases}
%     \frac{1}{N} \sum_{i=1}^N r_o(u, e_m^i),  & {\rm for}~t<n\\
%     r_o(u, e_n),  & {\rm for}~t=n
%     \end{cases}
% \end{equation}
\begin{equation}
r_s(s_t, a_t) =
    \begin{cases}
    \frac{1}{N} \sum_{i=1}^N r_o(u, e^{(i)}),  & {\rm for}~t<n\\
    r_o(u, e_n),  & {\rm for}~t=n
    \end{cases}
\end{equation}
In our approach, the scorer $\pi_s$ is the agent trained via SFT, ensuring its full capability of executing the required task.

\subsection{Iterative Agent Optimization}
\label{sec:training}

Agent tasks typically involve long action %task 
sequences and large decision spaces. 
Suppose we have a base agent $\pi_{\theta}$ trained through SFT. 
Given an instruction $u$, the agent interacts with the environment to produce a trajectory $e = (u, a_1, o_1, ..., a_n, o_n)$. 
% The agent might make an error action $a_t$ at step $t$. In order to correct this step, a straightforward approach would be to use reinforcement learning methods PPO~\cite{schulman2017proximal} to optimize the action at step $t$.
If the agent makes an error action $a_t$ at step $t$, a straightforward approach would be to use reinforcement learning methods like proximal policy optimization (PPO, \citealp{schulman2017proximal}) to optimize the action at step $t$.
However, applying online reinforcement learning directly to the LLM agent may cause practical issues such as instability~\citep{shen2023large, rafailov2024direct}. 
% Therefore, we perform offline learning on the contrastive action pairs data instead.
To address this issue, we perform offline learning on the contrastive action pairs data instead, which ensures stability.

\paragraph{Step-wise Trajectory Construction}

To generate contrastive action pairs data, we allow the base agent $\pi_\theta$ to explore on the expert trajectory. This approach has two benefits:
Firstly, upon identifying an incorrect action by the agent, we can easily acquire a correct action for contrastive learning purposes. 
% Second, we can avoid arbitrary exploration by the agent, resulting in a more meaningful trajectory.
Secondly, it prevents arbitrary exploration by the agent, thereby yielding a more informative trajectory.
For the task instruction $u$ with expert trajectory $e_n = (u, a_1, ..., o_{n-1}, a_n)$, we use the first $t-1$ steps $(u, a_1, ..., a_{t-1}, o_{t-1})$ as historical trajectory $e_{t-1}$.
The agent then predict the actions from step $t$ to get the trajectory:
\begin{equation}
    e_{t:m} = (\hat{a}_t, \hat{o}_t, ..., \hat{a}_m, \hat{o}_m),
\end{equation}
The rewards for $a_t$ and $\hat{a}_t$ are $r_s(s_t, a_t)$ and $r_s(s_t, \hat{a}_t)$, respectively.
We use a threshold $\tau$ to filter actions. If the reward of $\hat{a}_t$ is lower than that of $a_t$ by a margin greater than $\tau$, and the outcome reward of $\hat{e}_m$ is lower than that of $e_n$, we consider the agent to have made a mistake at step $t$. 
% And good-bad action pairs $a_w \succ a_l | u, e_{t-1} $ is constructed.
% And we contrast the subsequent trajectory from that step $e_w \succ e_l | u, e_{t-1} $
We then contrast the subsequent trajectory from that step $e^w_{t:n} \succ e^l_{t:m}\ \vert\ e_{t-1}$.
Here, $e^w$ and $e^l$ represent win/lose trajectories with higher and lower rewards. 
% Here, $w$ and $l$ represents win and lose respectively.
We perform exploration across the entire expert trajectory set and obtain the contrastive action dataset $\mathcal D_s = \Big\{(e_{t-1}, e^w_{t:n}, e^l_{t:m})^{(i)}\Big\}_{i=1}^{|\mathcal D_s|}$.
Additionally, we construct a contrastive trajectory dataset $\mathcal D_t = \Big\{(u, e^w_n, e^l_m)^{(i)}\Big\}_{i=1}^{|\mathcal D_t|}$ based on the outcome reward.

\paragraph{Mixture Trajectory Optimization}
During this phase, the agent policy undergoes updates through three loss components: outcome-DPO loss, step-DPO loss, and SFT loss.
Initially, to facilitate agent's learning from incorrect trajectories, we compute the outcome-DPO loss using the contrastive trajectory dataset:
\begin{equation}
\resizebox{1.0\hsize}{!}{$
\begin{aligned}
\mathcal L_{\mathrm{o\mbox{-}DPO}}= - \mathbb E_{(u, e^w_n, e^l_m)\sim \mathcal D_t}\bigg[\log\sigma(\beta\log\frac{\pi_\theta(e^w_n|u)}{\pi_{ref}(e^w_n|u)} \\
- \beta\log\frac{\pi_\theta(e^l_m|u)}{\pi_{ref}(e^l_m|u)} )\bigg],
\end{aligned}
$}
\end{equation}
% \begin{align*}
%     \mathcal L_{\mathrm{o-DPO}}= - \mathbb E_{(u, e_w, e_l)\sim \mathcal D_t}\bigg[\log\sigma(\beta\log\frac{\pi_\theta(e_w|u)}{\pi_{ref}(e_w|u)} \\
% - \beta\log\frac{\pi_\theta(e_l|u)}{\pi_{ref}(e_l|u)} )\bigg],
% \end{align*}
Next, the step-DPO loss imparts process-level supervision. Suppose the agent makes an error at step $t$, we have the agent performing a comparison for the subsequent trajectory, which is calculated as:
\begin{equation}
\resizebox{1.0\hsize}{!}{$
\begin{aligned}
\mathcal L_{\mathrm{s\mbox{-}DPO}}= - \mathbb E_{(e_{t-1}, e^w_{t:n},e^l_{t:m})\sim \mathcal D_s}\bigg[\log\sigma(\beta\log\frac{\pi_\theta(e^w_{t:n}|e_{t-1})}{\pi_{ref}(e^w_{t:n}|e_{t-1})} &\\
- \beta\log\frac{\pi_\theta(e^l_{t:m}|e_{t-1})}{\pi_{ref}(e^l_{t:m}|e_{t-1})} )\bigg]&,
\end{aligned}
$}
\end{equation}
% \begin{align*}
% & \mathcal{L}_{\mathrm{s-DPO}} = 
% \\&- \mathbb{E}_{(e_{t-1}, e_w, e_l)\sim \mathcal{D}_s} \bigg[ \log\sigma\bigg(\beta \log\frac{\pi_\theta(e_w|(e_{t-1})}{\pi_{ref}(e_w|(e_{t-1}))} \\
% &\quad - \beta \log\frac{\pi_\theta(e_l|(e_{t-1})}{\pi_{ref}(e_l|(e_{t-1}))} \bigg)\bigg]
% \end{align*}
% The o-DPO or s-DPO loss aims to increase the relative margin between the success and error pairs, enabling the agent to effectively learn from exloration failures. 
As demonstrated by \citet{yuan2024advancing}, DPO only optimizes the relative differences between chosen and rejected data, neglecting the absolute magnitudes of the rewards. This oversight can be problematic in agent tasks where the space of correct actions is significantly narrower than that of incorrect ones. To mitigate this issue, we add the SFT loss, aiming to directly increase the likelihood of the success trajectory:
% , maintaining the agent's performance during the inference:
\begin{equation}
\begin{aligned}
\mathcal L_{\mathrm{SFT}}= - \mathbb E_{(u, e^w_n, e^l_m)\sim \mathcal D_t}\bigg[\log \pi_\theta(e^w_n|u)\bigg],
\end{aligned}
\end{equation}
The final loss combines DPO and SFT losses:
\begin{equation}
    \mathcal L = \mathcal L_{\rm o\mbox{-}DPO} + \mathcal L_{\rm s\mbox{-}DPO} + \mathcal L_{\rm SFT}
\end{equation}

To further refine the agent's performance post-optimization, we employ the updated agent as the new base agent to continue collecting contrastive action pairs data for additional training. This iterative process is maintained until reaching the predetermined iteration limit.

\section{Experiments}
% In this section, we conduct extensive experiments to validate the effectiveness of \ourmethod{}. 
% Our method demonstrates superior performance compared to baselines across three datasets. The analysis further showcases the efficiency of our method.

\subsection{Experiment Settings}

\paragraph{Datasets} 
% We conduct experiments on three representative agent datasets, WebShop for web navigation, ALFWorld for embodied agent tasks and Intercode-SQL for SQL database querying. 
% Both ALFWorld and HotpotQA environments provide binary rewards indicating whether the task is completed, while WebShop provides dense reward ranging from 0 to 1.
% Both WebShop and Intercode-SQL provide dense reward ranging from 0 to 1, while ALFWorld provide binary rewards indicating whether the task is completed.

\begin{table}[t]
    \centering
    \resizebox{\linewidth}{!}{
    \begin{tabular}{l c c c c}
    \toprule
    \textbf{Dataset}   & \textbf{Train} & \textbf{Test} & \textbf{Action Space} & \textbf{Max Turns}\\
    \midrule
    WebShop & 1624 & 200 & 8 & 10\\
    ALFWorld & 2851 & 274 & 13 & 20\\
    InterCodeSQL & 1500 & 200 & $\infty$~(SQL) & 10\\
    \bottomrule
    \end{tabular}
    }
    \caption{Statistics overview of tested datasets. "Max Turns" refers to the maximum number of interactions in the expert trajectory.}
    \label{tab:dataset}
\end{table}

% We conduct experiments on three representative agent datasets to valid the effectiveness of \ourmethod{}: 
% WebShop~\citep{yao2022webshop} for web navigation, ALFWorld~\citep{shridhar2020alfworld} for embodied agent tasks and Intercode-SQL~\citep{yang2024intercode} for SQL database querying. 
% WebShop~\citep{yao2022webshop} is a simulated ecommerce website environment with 1.8 million real-world products and crowd-sourced text instructions. The agent needs to complete a series of actions, including searching, selecting products, choosing attributes, and finally make a purchase through 'search[]' or 'click[]'. WebShop will score the results from 0 to 1 based on how well the provide attributes match the requirements.
% ALFWorld~\citep{shridhar2020alfworld} is a household task where the agent needs to interact with objects in the text-described environment based on given task requirements, such as "put the apple in the refrigerator". Within given steps, the environment provides feedback on whether the agent successfully completes the task.
% InterCodeSQL~\citep{yang2024intercode} provides a database built on MySQL. The environment determines task completion via an exact match of the agent's execution output against the gold command, where 1 is awarded only if all components match.
We evaluate our method on three representative agent datasets: \textbf{WebShop}~\citep{yao2022webshop} for web navigation, \textbf{InterCodeSQL}~\citep{yang2024intercode} for SQL database querying, and \textbf{ALFWorld} for embodied agent tasks. Both WebShop and InterCodeSQL provide a dense reward scale from 0 to 1 to gauge task completion, while ALFWorld only provides a binary reward to indicate whether the task is completed. 
% We employ the \textbf{average reward} as the evaluation metric for all three tasks. 
We employ the \textbf{average reward} as the evaluation metric for all tasks. 

To collect training expert trajectories, we prompt GPT-4 to interact with the environment in ReAct pattern. We then filter the results based on the final outcome rewards to retain only the correct trajectories. Please refer to Appendix~\ref{appendix:collection} for more details.
The statistical information of the dataset is summarized in Table~\ref{tab:dataset}, and more details can be found in Appendix~\ref{appendix:datasets}. 
Note the ALFWorld test set is divided into 140 seen cases and 134 unseen cases, evaluating the agents' in-domain and out-of-domain proficiencies, respectively.

\paragraph{Implementation Details}
We utilize Llama-2-7B~\citep{touvron2023llama} as the base model to train LLM agents. 
The training epoch is 3 and with a batch size of 48. The AdamW optimizer~\citep{loshchilov2017decoupled} is employed, coupled with a cosine learning scheduler.
For step-level rewards acquisition via the scorer, we set the temperature to 1 and the number of samples $N$ to 5, promoting diversity in sampling. In the generation of contrastive action pairs, the base agent's temperature is fixed at 0, while the filtering threshold $\tau$ is adjusted to 0.5 for ALFWorld, 0.01 for WebShop and 0.1 for InterCodeSQL. All the generations are carried using \textit{vllm}~\citep{kwon2023efficient}.
During the mixture trajectory optimization phase, we search for the learning rate from 1e-5 to 5e-5, and $\beta$ for the DPO loss from 0.1 to 0.5. 
The iteration cap is set to 4. 
All experiments are conducted on a suite of 8 NVIDIA A100 80G GPUs.
% We have also included an analysis on training efficiency. Please refer to Appendix~\ref{appendix:training efficiency} for more details.

\paragraph{Baselines}
% \begin{table}[t]
% \centering
% \tabcolsep=4pt
% \resizebox{\linewidth}{!}{
% \begin{tabular}{lcccc}
% \toprule
% \textbf{Method} & \textbf{Exploration} & \textbf{Process Supervision} & \textbf{Robustness} & \textbf{} \\
% \midrule
% SFT & \xmark & \xmark & \cmark & \cmark \\
% RFT & \cmark & \xmark & \cmark & \cmark \\
% Online RL & \cmark & \cmark & \xmark & \xmark \\
% Offline RL & \xmark & \cmark & \cmark & \xmark \\
% \midrule
% \method{} & \cmark & \cmark & \cmark & \cmark \\
% \bottomrule
% \end{tabular}
% }
% \caption{
% \method{} vs alternatives: \method{} can leverage exploration failures to optimize the policy with high computational efficiency and robustness.
% }
% \label{tab:compare}
% \end{table}

We evaluate \ourmethod{} against three types of baselines: prompt-based, outcome refinement, and process refinement methods.
For prompt-based methods, we compare the efficacy of GPT-4~\citep{achiam2023gpt}, GPT-3.5-turbo~\citep{ouyang2022training}, and the untrained Llama-2-7B-Chat~\citep{touvron2023llama} utilizing ReAct prompting paradigm. These baselines are tested in a one-shot context.
Regarding outcome refinement methods, four tuning strategies are juxtaposed: (1) SFT~\citep{chen2023fireact} tunes the agent using solely expert trajectories, which is the base agent of other baselines; (2) PPO~\citep{schulman2017proximal} is a reinforcement learning (RL) technique that directly optimizes the agents to maximize the outcome reward; (3) RFT~(Rejection sampling Fine-Tuning)~\citep{yuan2023scaling} augments the expert trajectory dataset with successful trajectories, subsequently training the agent on the enriched dataset; and (4) ETO~\citep{song2024trial} contrasts success and failure trajectories via DPO~\citep{rafailov2024direct}.
For process refinement methods, we compare the Step-PPO method, which optimizes the agents to maximize the step-level process reward.

\subsection{Results} \label{sec:results}
% As shown in Table~\ref{tab:main results}, our \ourmethod{} framework achieves the highest performance compared to other baselines. Specifically, \ourmethod{} outperforms the previous state-of-the-art outcome refinement method ETO by 5.8\%, 7.2\%, 2.5\% and 3.2\% on WebShop, InterCodeSQL, ALFWorld~(seen), and AFLWorld~(unseen) respectively, with an average improvement of 4.5\%. This demonstrate the superiority of incorporating process supervision in enhancing agent performance. We hypothesize that ETO which only provides outcome supervision offers a significant weaker signal. This weaker signal is more challenging for the model to interpret and utilize effectively, making it harder to guide the model towards a successful policy. Consequently, this may result in the undesired penalization of correct actions. 

% 平均action的reward更重要，相同avg step reward的提升反映在最终的reward就更明显
% We can also observe that the performance improvement is more significant in WebShop compared to ALFWord. The difference may be related to the importance of each action. In WebShop, task trajectories are relatively shorter and each action has amore substantial impact on the final reward. While ALFWorld features longer task trajectories where many actions, such as checking rooms or moving around, have a smaller influence on the final reward. Therefore, the addition of process supervision in \ourmethod{} results in greater improvements in WebShop.

\begin{table*}[ht]
    \centering
    \resizebox{0.95\linewidth}{!}{
    \begin{tabular}{l l c c c c c}
    \toprule
    \multirow{2}{*}{\textbf{Paradigm}}& \multirow{2}{*}{\textbf{Models}} & \multirow{2}{*}{\textbf{WebShop}} & \multirow{2}{*}{\textbf{InterCodeSQL}} & \multicolumn{2}{c}{\textbf{ALFWorld}} & \multirow{2}{*}{\textbf{Average}}\\
    \cmidrule(l){5-6}
         & & & & Seen & Unseen \\
    \midrule
    \multirow{3}{*}{Prompt-based}& GPT-4~\citep{achiam2023gpt} & 63.2 & 38.5 & 42.9 & 38.1 & 45.7\\
    & GPT-3.5-Turbo~\citep{ouyang2022training} & 62.4 & 37.8 & 7.9 & 10.5 & 29.7\\
    & Llama-2-7B~\citep{touvron2023llama} & 17.9 & 4.0 & 0.0 & 0.0 & 5.5\\
    % & EURUS-7B \\
    \midrule
    \multirow{4}{*}{Outcome Refinement}
    & Llama-2-7B + SFT~\citep{chen2023fireact} & 60.2 & 54.9 & 60.0 & 67.2 & 60.6 \\
    & Llama-2-7B + PPO~\citep{schulman2017proximal} & 64.2 & 52.4 & 22.1 & 29.1 & 42.0 \\
    & Llama-2-7B + RFT~\citep{yuan2023scaling} & 63.6 & 56.3 & 62.9 & 66.4 & 62.3 \\
    & Llama-2-7B + ETO~\citep{song2024trial} & 67.4 & 57.2 & 68.6 & 72.4 & 66.4 \\ 
    \midrule
    \multirow{2}{*}{Process Refinement} & Llama-2-7B + Step-PPO & 64.0 & 60.2 & 65.7 & 69.4 & 64.8\\
    & \textbf{Llama-2-7B + \ourmethod{}~(ours)} & \textbf{71.3} & \textbf{61.3} & \textbf{70.3} & \textbf{74.7} &  \textbf{69.4}\\
    \bottomrule
    \end{tabular}
    }
    \caption{Performance of different methods on three agent datasets. \ourmethod{} shows superiority over prompt-based and outcome refinement methods. For ETO and \ourmethod{}, we report the best performance across all iterations.}
    \label{tab:main results}
\end{table*}

Table~\ref{tab:main results} illustrates that, in comparison to outcome refinement and process refinement methods, both open-source and proprietary models under prompt-based methods perform significantly worse. This discrepancy is particularly evident with the untrained Llama-2-7B, which struggles to complete the InterCodeSQL and ALFWorld tasks. However, after training with our \ourmethod{} method, there is a remarkable increase in the average reward from 5.5 to 69.4, surpassing the best performance of GPT-4.
Regarding outcome refinement baselines, our method outperforms the previous state-of-the-art (SOTA) method ETO by margins of 5.8\%, 7.2\%, 2.5\% and 3.2\% on WebShop, InterCodeSQL, ALFWorld~(seen), and AFLWorld~(unseen) respectively, with an average improvement of 4.5\%. This underscores the superiority of integrating process supervision in enhancing agent performance.
As for process refinement baselines, while Step-PPO performs well on InterCodeSQL, surpassing both prompt-based and outcome refinement baselines, its instability within RL optimization procedures results in poor performance on the other two tasks. In contrast, \ourmethod{} significantly enhances agent performance, outperforming all baselines across the three complex interactive agent tasks. We also present case studies to delineat the task-solving trajectories of our method in Appendix~\ref{appendix:case study}.
Moreover, \ourmethod{} showcases robustness on the ALFWorld unseen task, affirming its generalization capabilities.
We have also included an analysis on training efficiency. Please refer to Appendix~\ref{appendix:training efficiency} for more details.

\section{Analysis}\label{sec:analysis}
\subsection{Different Base Models}
% llama2-13b, mistral, llama3
To further substantiate the efficacy of our method, we conduct validations across a variety of base models. We select Mistral-7B~\citep{jiang2023mistral}, Llama-2-13B~\citep{touvron2023llama} and Llama-3-8B~\citep{meta2024introducing} as our base LLMs, employing WebShop and InterCodeSQL as evaluation datasets. We juxtapose the performance of \ourmethod{} with that of ETO and SFT. The comparative results are summarized in Table~\ref{tab:base model}.
\ourmethod{} consistently outperforms ETO and SFT across all models and datasets. Notably, on the Mistral model, where SFT performance is relatively poor, our method realizes a significant improvement, demonstrating that our approach can effectively enhance the performance of weaker models.
Furthermore, we observe that on the WebShop task, Llama-2-13B achieves the best performance after SFT and maintains its leading position after \ourmethod{}. Similarly, Llama-3-8B shows superior performance on the InterCodeSQL task. This pattern indicates that base agents with higher initial performance are prone to achieve more pronounced final performance post-\ourmethod{} training.

\begin{table}[t]
    \centering
    \resizebox{\linewidth}{!}{
    \begin{tabular}{l | c | c | c}
    \toprule
    \textbf{Base LLM}  &  \textbf{Setting} & \textbf{WebShop} & \textbf{InterCodeSQL}\\ \midrule
    \multirow{3}{*}{Mistral-7B}    & SFT & 58.5 & 50.0\\
    & ETO & 66.2 & 54.3\\
    & \ourmethod{} & \textbf{69.6} & \textbf{58.9} \\ \midrule
    \multirow{3}{*}{Llama-2-13B} & SFT & 62.2 & 59.3\\
    & ETO & 68.9 & 61.5\\
    & \ourmethod{} & \textbf{72.2} & \textbf{64.5}\\ \midrule
    \multirow{3}{*}{Llama-3-8B} & SFT & 61.2 & 63.4\\
    & ETO & 66.2 & 65.8\\
    % & \ourmethod{} & \textbf{68.8} &  \textbf{68.1}\\
    & \ourmethod{} & \textbf{72.0} &  \textbf{68.1}\\
    \bottomrule
    \end{tabular}
    }
    \caption{The performance of different base LLMs on WebShop and InterCodeSQL.}
    \label{tab:base model}
\end{table}

\subsection{Ablation Study}
% % SFT loss, Step-Traj
We conduct ablation experiments on the training methods and iteration rounds for \ourmethod{}. For ALFWorld, we evaluate performance on the unseen test set. 
As shown in Table~\ref{tab:ablation}, removing each module results in a clear drop in the agent's performance, underscoring the power of our method.
For the ablation on training methods, we discern that the removal of SFT loss engenders the most pronounced performance drop in the agent.
% We hypothesize that this is because SFT avoids the issue where DPO only optimizes the relative differences between chosen and rejected data, overlooking the absolute values of the rewards. 
Additionally, we find that removing the step-DPO loss induce a more substantial performance decline than that of removing the outcome-DPO loss, suggesting the necessity of process supervision. 

The iteration ablation results show that in the initial rounds of iteration, the agent continually refine its performance by learning from incorrect actions. 
% However, on ALFWorld, the performance in the fourth iteration decreases compared to the third iteration. 
However, excessive iterations can result in a decrease in performance.
This decline might be attributed to overfitting, a consequence of excessive exploration of the training set. 

\begin{table}[t]
    \centering
    \resizebox{\linewidth}{!}{
    \begin{tabular}{l | c  c  c} \toprule
    \textbf{Training Scheme}  &  \textbf{WebShop}  & \textbf{InterCodeSQL} & \textbf{ALFWorld} \\ \midrule
     w/o o-DPO & 70.2 & 59.3 & 72.4 \\
     w/o s-DPO & 66.4 & 58.0 & 70.2 \\
     w/o SFT & 61.8 & 31.7 & 64.9 \\ \midrule
     Iteration=1 & 63.6 & 56.6 & 68.7 \\
     Iteration=2 & 63.7 & 58.2 & 70.2 \\
     Iteration=3 & 68.2 & 59.2 & \textbf{74.7} \\
     Iteration=4 & \textbf{71.3} & \textbf{61.3} & 73.5 \\
     Iteration=5 & 68.1 & 57.9 & 71.4 \\ \bottomrule
    \end{tabular}
    }
    \caption{Ablation study on training methods and iterations.}
    \label{tab:ablation}
\end{table}

% \subsection{Success Trajectory Length}
% \subsection{Efficiency Analysis}
\subsection{Step Reward Estimation Quality}
The employment of a scorer agent to estimate process rewards may introduce some noise. To evaluate the accuracy of step rewards, we conduct an experimental analysis on WebShop. In WebShop, each action navigates to a new web page, and scoring rules are established to calculate the final reward for purchasing a product. ~\citet{ma2024agentboard} heuristically expands the product scoring rules to assign scores at different web pages, thereby scoring each action. This helps us evaluate the quality of two different actions taken from the same state. Please refer to Appendix~\ref{appendix:scoring} for more details.
We define accuracy as the ratio of our constructed contrastive action pairs' order that satisfy the scoring function introduced by ~\citet{ma2024agentboard}. We analyze the impact of using different LLM agents as scorers and varying the Monte Carlo sampling times on the accuracy of step reward estimation. When constructing the contrastive action pairs, we set the reward threshold $\tau$ as 0.35 for all base models.

\begin{figure}[t]
    \centering
    \includegraphics[width=0.45\textwidth]{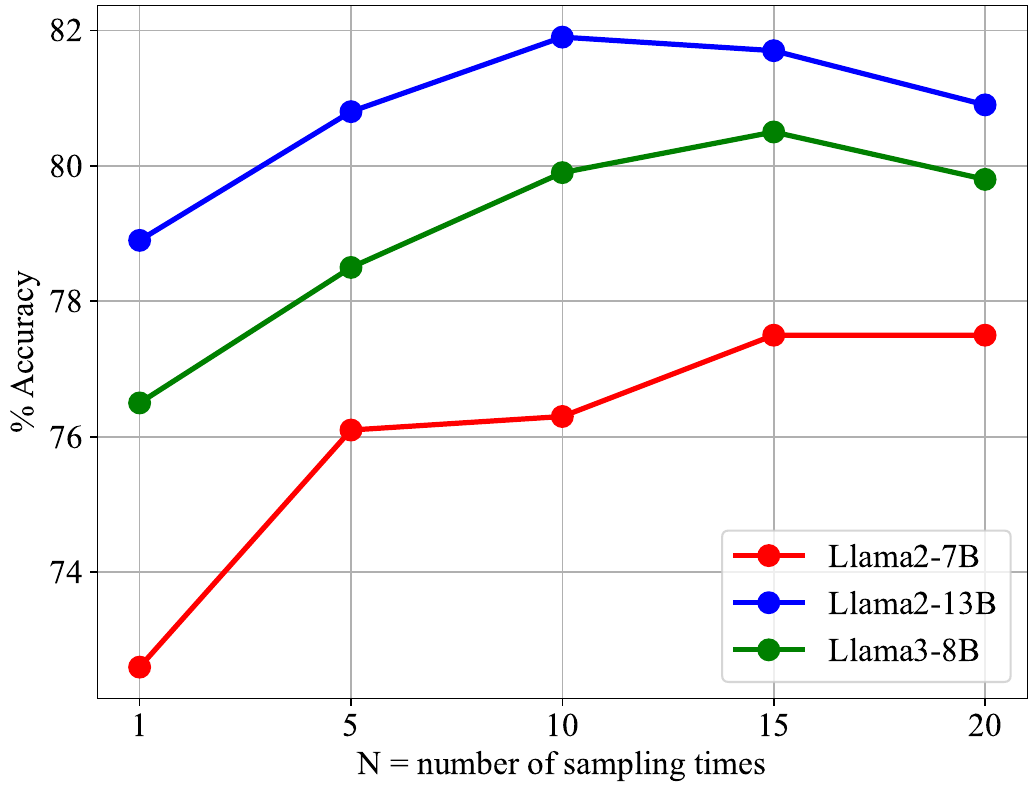}
    \caption{Step reward estimation quality on WebShop.}
    \label{fig:precision}
\end{figure}

Figure~\ref{fig:precision} illustrates that, despite inherent noise, the sampling approach yields satisfactory process reward estimations, achieving an accuracy of up to 82\% . The accuracy is influenced by the base model's performance on the task. For example, with the same sample count, Llama-2-13B achieves the highest quality in step reward estimation. 
This suggests that using a more powerful base model~(Table~\ref{tab:base model}) can improve the quality of step reward annotations. 
Additionally, the number of samples affects step reward estimation quality. Increasing samples can improve scoring accuracy but raise time costs. Despite the efficiency concerns with MC method, we can balance sample size and scoring accuracy. For WebShop, setting the sampling number $N=5$ achieves performance comparable to a larger sample size.
Without increasing inference time costs, \ourmethod{} achieves nearly a 6\% performance improvement at the expense of three times the ETO training duration.

\subsection{Average Reward Per Step}
\begin{figure}[t]
    \centering
    \includegraphics[width=0.48\textwidth]{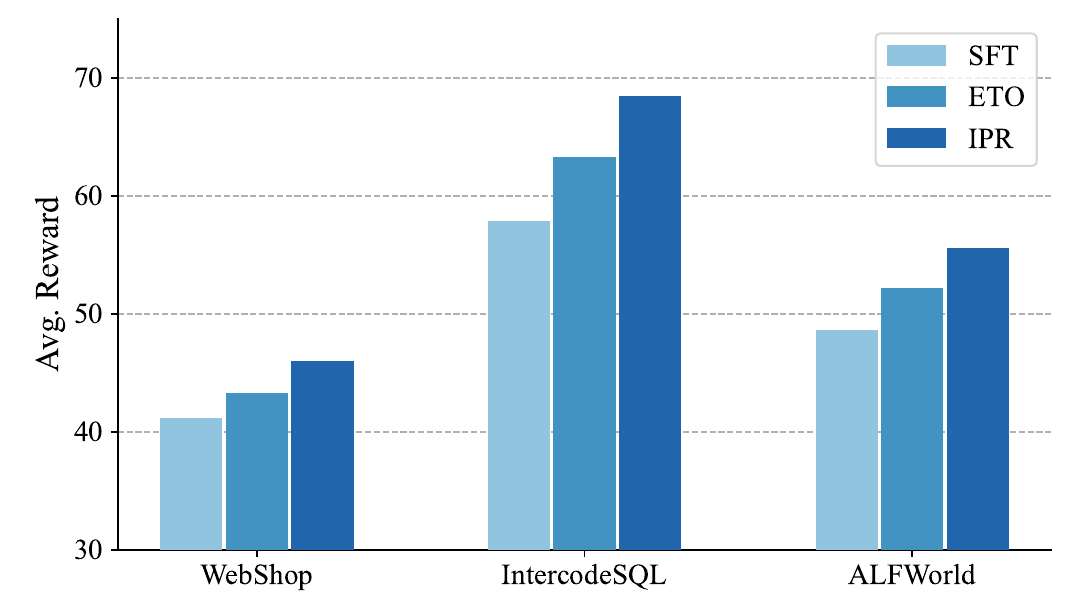}
    \caption{The average reward per step.}
    \label{fig:Average Reward}
\end{figure}

The purpose of \ourmethod{} is to provide process-level supervision to the agent, enabling it to take more accurate actions at each step. Here, we evaluate the changes in the average reward per step after training. The reward for each step is estimated according to the procedure in Section~\ref{sec:reward estimation}. 
We calculate the average rewards for all actions within each trajectory and then average these values across the entire test set. 
Figure~\ref{fig:Average Reward} illustrates the significant improvements in average step rewards achieved by our \ourmethod{} method compared to SFT and ETO across three tasks. 
% Compared to WebShop, which has shorter trajectories, \ourmethod{} achieves greater improvements on ALFWorld.
It can also be observed that for datasets where SFT training has a higher average step reward, such as InterCodeSQL, the improvement in step reward is even more pronounced.
These results underscore the superior performance of  \ourmethod{}, confirming its effectiveness in enhancing the accuracy and efficacy of agent actions.

% \subsection{Exploration of Step Reward Data}
\subsection{Exploration of Step Reward Modeling}

% 用蒙特卡洛采样的轨迹训练reward model
%In this section, 
% We further investigate the feasibility of constructing a step-level reward model to reduce the computational overhead of using the MC method in \ourmethod{} for reward estimation.
% Given the historical trajectory $e_{t-1}$ and the current action $a_t$, the reward model outputs a score as the step reward.
% Further, we utilize step reward data obtained from the MC method to train an environment's step reward model, thereby reducing the training time for new models within that environment.
% To reduce the training time for new models within that environment, here, we utilize step reward data obtained from the MC method to train an environment's step reward model.
Based on the step reward data we collected, we conduct further exploration and develop a step reward model, which can reduce the training time for new models within that environment.
Given the historical trajectory $e_{t-1}$ and the current action $a_t$, the reward model outputs a score as the step reward.
We conduct experiments on WebShop, using Llama-2-7B to build the reward model. 
We collect 70k actions generated by Llama-2-7B and Llama-2-13B as training data, with the step rewards estimated using the MC method. We train the reward model with MSE loss.
To evaluate the effectiveness of the reward model, we replace the scorer in Section~\ref{sec:reward estimation} with the reward model and compare the results against ETO (which does not use step rewards) and the MC method. 
% As shown in Table~\ref{tab:reward model}, although the results are lower than MC method, they still outperform ETO, indicating the capability of the step reward model.
As shown in Table~\ref{tab:reward model}, the reward model can enhance the performance of Llama-3-8B, even though its actions are not included in the training data. This indicates the generalization and robustness of the reward model. However, despite outperforming ETO, the results still fall short of the MC method. This may be attributed to the model's less accurate estimation of step rewards within the environment, suggesting the need for further improvement.
% Additionally, we conduct experiments on Llama-3-8B, whose actions are not included in the training data. The reward model could improve the agent's performance, demonstrating its generalization and robustness.
% The step reward model has other potential applications, such as guiding the agent's action decisions or refining the evaluation metrics for agent tasks. We leave these aspects in future work.

\begin{table}[t]
    \centering
    \resizebox{\linewidth}{!}{
    \begin{tabular}{l | c  c  c}
    \toprule
    \textbf{Models} & \textbf{No Reward} & \textbf{Reward Model} & \textbf{MC Method} \\ \midrule
     Llama-2-7B & 67.4 & 68.9 & 71.3 \\
     Llama-2-13B & 68.9 & 70.7 & 72.2 \\
     Llama-3-8B & 66.2 & 70.6 &  72.0\\
    \bottomrule
    \end{tabular}
    }
    \caption{The performance of different step reward acquisition methods.}
    \label{tab:reward model}
\end{table}

\section{Related Work}
\subsection{LLM as Agents}
% Agents need multi-dimensional skills to tackle interactive decision-making problems. 
% While reinforcement learning offers general solution, it struggles with sample efficiency and generalization problems~\citep{pourchot2018cem}. 
The emerging reasoning and instruction-following capabilities of LLMs~\citep{wei2022emergent} enable them to act as adept agents, particularly in zero-shot generalization across new tasks and problems~\cite{yao2022react, richardssignificant, wang2023voyager}. The key technique involves formulating prompts that furnish LLMs with instructions and context about the environment, thereby enabling them to generate executable actions and leverage external tools for complex task-solving~\citep{song2023restgpt, xie2023openagents}. To enhance the capabilities of open-source LLMs as agents, recent efforts have adopted fine-tuning methods~\cite{chen2023fireact, zeng2023agenttuning, yin2023lumos}. These methods enables agent learn from successful trajectories or utilize contrastive information with failed trajectories~\citep{song2024trial}. However, these approaches only leverage final outcome reward, with no studies to date investigating the integration of process information to improve agent performance.

\subsection{Step-level Process Supervision}
% Outcome-based methods only supervise the final result. 
In the resolution of complex tasks, even SOTA models may still make mistakes at intermediate steps.
To monitor the task completion process and avoid such errors, some approaches~\citep{uesato2022solving, lightman2023let} employ process-based methods which can provide 
% more precise feedback and specify the exact location of any errors that occur. 
step-level guidance.
To avoid the high cost of manually collecting process supervision, recent works~\citep{liu2023don, wang2023math, havrilla2024glore, wang2024multi} construct pseudo-labels, using the model's potential to complete the task given the previous steps as process labels. 
These methods~\citep{ma2023let, luong2024reft} use PPO to optimize the model but suffer from training efficiency and instability issues. Our approach, designed with mixture trajectory optimization, effectively enhances the agent's performance.
% However, these methods still require additional training of a reward model with pseudo-labels, which is used for PPO or answer re-ranking during inference. Our approach is the first to apply process rewards directly to agent tasks which does not require additional modules and results in more stable training outcomes.

\subsection{Self-Improvement}
% Despite the breakthroughs large language models~(LLMs) have achieved in many fields, training these models requires a substantial amont of human-labeld data or external model supervision~\citep{tao2024survey}. To address this issue, self-improvements have gained attention. These methods allow the model to autonomously acquire, refine, and learn from experiences generated by the model itself. 
To compensate for the scarcity of high-quality training data~\citep{tao2024survey}, self-improvement methods empower the model to autonomously acquire, refine, and learn from self-generated experiences.
% In the field of code generation, SelfEvolve~\cite{jiang2023selfevolve} and LDB~\citep{zheng2024judging} enable the model to produce code and learn from feedback obtained by executing the code in an interpreter. SPIN~\citep{chen2024self} generates its own training data from its previous iterations, refining its policy by discerning these self-generated responses from those obtained from human-annotated data. 
% For agent abilities, AutoAct~\citep{qiao2024autoact} introducing self-planning from scratch, focusing on an intrinsic self-learning process. In this process, agents enhance their abilities through recursive planning iterations with environment feedback.
% In the field of code generation, SelfEvolve~\cite{jiang2023selfevolve} and LDB~\citep{zheng2024judging} enable the model to produce code and learn from feedback obtained by executing the code in an interpreter.
Certain works~\citep{jiang2023selfevolve, singh2023beyond, zelikman2023self, chen2024self} focus on alignment, refining the model by discerning these self-generated responses from those obtained from human-annotated data. 
Others concentrate on LLM agents utilized for task-solving and interaction in dynamic environments. They enhance the agent's capabilities in planning~\citep{qiao2024autoact}, tool using~\citep{bousmalis2023robocat,zhu2024knowagent}, and communication~\citep{ulmer2024bootstrapping}.
These endeavors demonstrate that models can refine themselves through exploration in diverse domains. Our work aims to amplify this self-improvement process by providing fine-grained guidance.

\section{Conclusion}
In this paper, we present \ourmethod{}, a novel framework designed to elevate the capabilties of LLM agents in complex interaction tasks. Our approach integrates process-level supervision, enabling agents to learn from contrast action pairs. 
To provide fine-grained guidance in environments where only outcome rewards are available, we use the MC method to automatically calculate step rewards.
By employing iterative agent optimization, \ourmethod{} provides an effective way to optimize agent decision-making trajectories. 
Experiments on three benchmarks demonstrate that our framework consistently outperforms existing baselines. 
Subsequent analyses validate the efficacy of each part of the framework and action efficiency. 
We believe the \ourmethod{} framework can serve as a potent tool for enhancing agent performance at the action level, thereby catalyzing future progress in intelligent agent development.

\section*{Limitations}
Despite achieving the best performance compared to other baselines, it is important to acknowledge several limitations of this work. 
1) Our method provides fine-grained supervision for the agent's self-improvement process. However due to limited training data, which is a quite common scenario, iterative preference learning on self-generated samples can lead to overfitting. Future work could explore the augmentation of training tasks using GPT-4 to mitigate this issue. 
% 2) We identify error actions based on step rewards to construct contrastive data but do not utilize the numerical values of the step rewards, which could indicate the degree of error at each step. Adopting a curriculum learning approach~\citep{wang2021survey}, where more erroneous actions are corrected first before addressing less erroneous ones, might further enhance agent performance. 
% 2) While our method identifies error actions through step rewards to create contrastive datasets, it does not fully exploit the potential of these rewards. The numerical values of step rewards could indicate the severity of errors at each step. Adopting the curriculum learning approach~\citep{wang2021survey}, where more erroneous actions are corrected first before addressing less erroneous ones, might further enhance agent performance.
2) Our method only explores identifying error actions and creating contrastive datasets through step rewards. However, it does not fully exploit the potential of these rewards. The numerical values of step rewards could indicate the severity of errors at each step. For instance, adopting the curriculum learning approach~\citep{wang2021survey}, where more severe errors are corrected first before addressing less significant ones, might further enhance agent performance.
% 3) Our scorer employs a model trained through SFT, with its parameters fixed. However, it can also be continuously enhanced with the improvement of the base agent.
3) Our step reward model is only trained on a single agent task, which affects its generalizability across different tasks. Future work could develop a general agent step reward model applicable to various tasks.

\section*{Acknowledgement}
We thank the anonymous reviewers for their helpful comments on this paper. This work was partially supported by National Natural Science Foundation of China (No. 62476010). 

% comment for arxiv

% \section*{Ethics Statement}
% This work fully complies with the ACL Ethics Policy.
% We declare that there are no ethical issues in this paper, to the best of our knowledge.

\bibliography{acl_latex}

\begin{thebibliography}{46}
\providecommand{\natexlab}[1]{#1}

\bibitem[{Achiam et~al.(2023)Achiam, Adler, Agarwal, Ahmad, Akkaya, Aleman, Almeida, Altenschmidt, Altman, Anadkat et~al.}]{achiam2023gpt}
Josh Achiam, Steven Adler, Sandhini Agarwal, Lama Ahmad, Ilge Akkaya, Florencia~Leoni Aleman, Diogo Almeida, Janko Altenschmidt, Sam Altman, Shyamal Anadkat, et~al. 2023.
\newblock Gpt-4 technical report.
\newblock \emph{arXiv preprint arXiv:2303.08774}.

\bibitem[{Bousmalis et~al.(2023)Bousmalis, Vezzani, Rao, Devin, Lee, Villalonga, Davchev, Zhou, Gupta, Raju et~al.}]{bousmalis2023robocat}
Konstantinos Bousmalis, Giulia Vezzani, Dushyant Rao, Coline~Manon Devin, Alex~X Lee, Maria~Bauza Villalonga, Todor Davchev, Yuxiang Zhou, Agrim Gupta, Akhil Raju, et~al. 2023.
\newblock Robocat: A self-improving generalist agent for robotic manipulation.
\newblock \emph{Transactions on Machine Learning Research}.

\bibitem[{Chen et~al.(2023)Chen, Shu, Shareghi, Collier, Narasimhan, and Yao}]{chen2023fireact}
Baian Chen, Chang Shu, Ehsan Shareghi, Nigel Collier, Karthik Narasimhan, and Shunyu Yao. 2023.
\newblock Fireact: Toward language agent fine-tuning.
\newblock \emph{arXiv preprint arXiv:2310.05915}.

\bibitem[{Chen et~al.(2024)Chen, Deng, Yuan, Ji, and Gu}]{chen2024self}
Zixiang Chen, Yihe Deng, Huizhuo Yuan, Kaixuan Ji, and Quanquan Gu. 2024.
\newblock Self-play fine-tuning converts weak language models to strong language models.
\newblock \emph{arXiv preprint arXiv:2401.01335}.

\bibitem[{Havrilla et~al.(2024)Havrilla, Raparthy, Nalmpantis, Dwivedi-Yu, Zhuravinskyi, Hambro, and Railneau}]{havrilla2024glore}
Alex Havrilla, Sharath Raparthy, Christoforus Nalmpantis, Jane Dwivedi-Yu, Maksym Zhuravinskyi, Eric Hambro, and Roberta Railneau. 2024.
\newblock Glore: When, where, and how to improve llm reasoning via global and local refinements.
\newblock \emph{arXiv preprint arXiv:2402.10963}.

\bibitem[{Jiang et~al.(2023{\natexlab{a}})Jiang, Sablayrolles, Mensch, Bamford, Chaplot, Casas, Bressand, Lengyel, Lample, Saulnier et~al.}]{jiang2023mistral}
Albert~Q Jiang, Alexandre Sablayrolles, Arthur Mensch, Chris Bamford, Devendra~Singh Chaplot, Diego de~las Casas, Florian Bressand, Gianna Lengyel, Guillaume Lample, Lucile Saulnier, et~al. 2023{\natexlab{a}}.
\newblock Mistral 7b.
\newblock \emph{arXiv preprint arXiv:2310.06825}.

\bibitem[{Jiang et~al.(2023{\natexlab{b}})Jiang, Wang, and Wang}]{jiang2023selfevolve}
Shuyang Jiang, Yuhao Wang, and Yu~Wang. 2023{\natexlab{b}}.
\newblock Selfevolve: A code evolution framework via large language models.
\newblock \emph{arXiv preprint arXiv:2306.02907}.

\bibitem[{Kwon et~al.(2023)Kwon, Li, Zhuang, Sheng, Zheng, Yu, Gonzalez, Zhang, and Stoica}]{kwon2023efficient}
Woosuk Kwon, Zhuohan Li, Siyuan Zhuang, Ying Sheng, Lianmin Zheng, Cody~Hao Yu, Joseph~E. Gonzalez, Hao Zhang, and Ion Stoica. 2023.
\newblock Efficient memory management for large language model serving with pagedattention.
\newblock In \emph{Proceedings of the ACM SIGOPS 29th Symposium on Operating Systems Principles}.

\bibitem[{Lightman et~al.(2023)Lightman, Kosaraju, Burda, Edwards, Baker, Lee, Leike, Schulman, Sutskever, and Cobbe}]{lightman2023let}
Hunter Lightman, Vineet Kosaraju, Yura Burda, Harri Edwards, Bowen Baker, Teddy Lee, Jan Leike, John Schulman, Ilya Sutskever, and Karl Cobbe. 2023.
\newblock Let's verify step by step.
\newblock \emph{arXiv preprint arXiv:2305.20050}.

\bibitem[{Liu et~al.(2023)Liu, Cohen, Pasunuru, Choi, Hajishirzi, and Celikyilmaz}]{liu2023don}
Jiacheng Liu, Andrew Cohen, Ramakanth Pasunuru, Yejin Choi, Hannaneh Hajishirzi, and Asli Celikyilmaz. 2023.
\newblock Don't throw away your value model! making ppo even better via value-guided monte-carlo tree search decoding.
\newblock \emph{arXiv e-prints}, pages arXiv--2309.

\bibitem[{Loshchilov and Hutter(2017)}]{loshchilov2017decoupled}
Ilya Loshchilov and Frank Hutter. 2017.
\newblock Decoupled weight decay regularization.
\newblock \emph{arXiv preprint arXiv:1711.05101}.

\bibitem[{Luong et~al.(2024)Luong, Zhang, Jie, Sun, Jin, and Li}]{luong2024reft}
Trung~Quoc Luong, Xinbo Zhang, Zhanming Jie, Peng Sun, Xiaoran Jin, and Hang Li. 2024.
\newblock Reft: Reasoning with reinforced fine-tuning.
\newblock \emph{arXiv preprint arXiv:2401.08967}.

\bibitem[{Ma et~al.(2024)Ma, Zhang, Zhu, Yang, Yang, Jin, Lan, Kong, and He}]{ma2024agentboard}
Chang Ma, Junlei Zhang, Zhihao Zhu, Cheng Yang, Yujiu Yang, Yaohui Jin, Zhenzhong Lan, Lingpeng Kong, and Junxian He. 2024.
\newblock Agentboard: An analytical evaluation board of multi-turn llm agents.
\newblock \emph{arXiv preprint arXiv:2401.13178}.

\bibitem[{Ma et~al.(2023)Ma, Zhou, Liu, Yuan, Liu, You, and Yang}]{ma2023let}
Qianli Ma, Haotian Zhou, Tingkai Liu, Jianbo Yuan, Pengfei Liu, Yang You, and Hongxia Yang. 2023.
\newblock Let's reward step by step: Step-level reward model as the navigators for reasoning.
\newblock \emph{arXiv preprint arXiv:2310.10080}.

\bibitem[{Meta(2024)}]{meta2024introducing}
AI~Meta. 2024.
\newblock Introducing meta llama 3: The most capable openly available llm to date.
\newblock \emph{Meta AI.}

\bibitem[{Ouyang et~al.(2022)Ouyang, Wu, Jiang, Almeida, Wainwright, Mishkin, Zhang, Agarwal, Slama, Ray et~al.}]{ouyang2022training}
Long Ouyang, Jeffrey Wu, Xu~Jiang, Diogo Almeida, Carroll Wainwright, Pamela Mishkin, Chong Zhang, Sandhini Agarwal, Katarina Slama, Alex Ray, et~al. 2022.
\newblock Training language models to follow instructions with human feedback.
\newblock \emph{Advances in neural information processing systems}, 35:27730--27744.

\bibitem[{Qiao et~al.(2024)Qiao, Zhang, Fang, Luo, Zhou, Jiang, Lv, and Chen}]{qiao2024autoact}
Shuofei Qiao, Ningyu Zhang, Runnan Fang, Yujie Luo, Wangchunshu Zhou, Yuchen~Eleanor Jiang, Chengfei Lv, and Huajun Chen. 2024.
\newblock Autoact: Automatic agent learning from scratch via self-planning.
\newblock \emph{arXiv preprint arXiv:2401.05268}.

\bibitem[{Rafailov et~al.(2024)Rafailov, Sharma, Mitchell, Manning, Ermon, and Finn}]{rafailov2024direct}
Rafael Rafailov, Archit Sharma, Eric Mitchell, Christopher~D Manning, Stefano Ermon, and Chelsea Finn. 2024.
\newblock Direct preference optimization: Your language model is secretly a reward model.
\newblock \emph{Advances in Neural Information Processing Systems}, 36.

\bibitem[{Richards(2023)}]{richardssignificant}
Toran~Bruce Richards. 2023.
\newblock Significant-gravitas/autogpt: An experimental open-source attempt to make gpt-4 fully autonomous.
\newblock \emph{URL https://github. com/Significant-Gravitas/AutoGPT}.

\bibitem[{Schulman et~al.(2017)Schulman, Wolski, Dhariwal, Radford, and Klimov}]{schulman2017proximal}
John Schulman, Filip Wolski, Prafulla Dhariwal, Alec Radford, and Oleg Klimov. 2017.
\newblock Proximal policy optimization algorithms.
\newblock \emph{arXiv preprint arXiv:1707.06347}.

\bibitem[{Shen et~al.(2023)Shen, Jin, Huang, Liu, Dong, Guo, Wu, Liu, and Xiong}]{shen2023large}
Tianhao Shen, Renren Jin, Yufei Huang, Chuang Liu, Weilong Dong, Zishan Guo, Xinwei Wu, Yan Liu, and Deyi Xiong. 2023.
\newblock Large language model alignment: A survey.
\newblock \emph{arXiv preprint arXiv:2309.15025}.

\bibitem[{Shinn et~al.(2024)Shinn, Cassano, Gopinath, Narasimhan, and Yao}]{shinn2024reflexion}
Noah Shinn, Federico Cassano, Ashwin Gopinath, Karthik Narasimhan, and Shunyu Yao. 2024.
\newblock Reflexion: Language agents with verbal reinforcement learning.
\newblock \emph{Advances in Neural Information Processing Systems}, 36.

\bibitem[{Shridhar et~al.(2020)Shridhar, Yuan, C{\^o}t{\'e}, Bisk, Trischler, and Hausknecht}]{shridhar2020alfworld}
Mohit Shridhar, Xingdi Yuan, Marc-Alexandre C{\^o}t{\'e}, Yonatan Bisk, Adam Trischler, and Matthew Hausknecht. 2020.
\newblock Alfworld: Aligning text and embodied environments for interactive learning.
\newblock \emph{arXiv preprint arXiv:2010.03768}.

\bibitem[{Singh et~al.(2023)Singh, Co-Reyes, Agarwal, Anand, Patil, Liu, Harrison, Lee, Xu, Parisi et~al.}]{singh2023beyond}
Avi Singh, John~D Co-Reyes, Rishabh Agarwal, Ankesh Anand, Piyush Patil, Peter~J Liu, James Harrison, Jaehoon Lee, Kelvin Xu, Aaron Parisi, et~al. 2023.
\newblock Beyond human data: Scaling self-training for problem-solving with language models.
\newblock \emph{arXiv preprint arXiv:2312.06585}.

\bibitem[{Song et~al.(2023)Song, Xiong, Zhu, Li, Wang, Tian, and Li}]{song2023restgpt}
Yifan Song, Weimin Xiong, Dawei Zhu, Cheng Li, Ke~Wang, Ye~Tian, and Sujian Li. 2023.
\newblock Restgpt: Connecting large language models with real-world applications via restful apis.
\newblock \emph{arXiv preprint arXiv:2306.06624}.

\bibitem[{Song et~al.(2024)Song, Yin, Yue, Huang, Li, and Lin}]{song2024trial}
Yifan Song, Da~Yin, Xiang Yue, Jie Huang, Sujian Li, and Bill~Yuchen Lin. 2024.
\newblock Trial and error: Exploration-based trajectory optimization for llm agents.
\newblock \emph{arXiv preprint arXiv:2403.02502}.

\bibitem[{Tao et~al.(2024)Tao, Lin, Chen, Li, Wu, Li, Jin, Huang, Tao, and Zhou}]{tao2024survey}
Zhengwei Tao, Ting-En Lin, Xiancai Chen, Hangyu Li, Yuchuan Wu, Yongbin Li, Zhi Jin, Fei Huang, Dacheng Tao, and Jingren Zhou. 2024.
\newblock A survey on self-evolution of large language models.
\newblock \emph{arXiv preprint arXiv:2404.14387}.

\bibitem[{Touvron et~al.(2023)Touvron, Martin, Stone, Albert, Almahairi, Babaei, Bashlykov, Batra, Bhargava, Bhosale et~al.}]{touvron2023llama}
Hugo Touvron, Louis Martin, Kevin Stone, Peter Albert, Amjad Almahairi, Yasmine Babaei, Nikolay Bashlykov, Soumya Batra, Prajjwal Bhargava, Shruti Bhosale, et~al. 2023.
\newblock Llama 2: Open foundation and fine-tuned chat models.
\newblock \emph{arXiv preprint arXiv:2307.09288}.

\bibitem[{Uesato et~al.(2022)Uesato, Kushman, Kumar, Song, Siegel, Wang, Creswell, Irving, and Higgins}]{uesato2022solving}
Jonathan Uesato, Nate Kushman, Ramana Kumar, Francis Song, Noah Siegel, Lisa Wang, Antonia Creswell, Geoffrey Irving, and Irina Higgins. 2022.
\newblock Solving math word problems with process-and outcome-based feedback.
\newblock \emph{arXiv preprint arXiv:2211.14275}.

\bibitem[{Ulmer et~al.(2024)Ulmer, Mansimov, Lin, Sun, Gao, and Zhang}]{ulmer2024bootstrapping}
Dennis Ulmer, Elman Mansimov, Kaixiang Lin, Justin Sun, Xibin Gao, and Yi~Zhang. 2024.
\newblock Bootstrapping llm-based task-oriented dialogue agents via self-talk.
\newblock \emph{arXiv preprint arXiv:2401.05033}.

\bibitem[{Wang et~al.(2023{\natexlab{a}})Wang, Xie, Jiang, Mandlekar, Xiao, Zhu, Fan, and Anandkumar}]{wang2023voyager}
Guanzhi Wang, Yuqi Xie, Yunfan Jiang, Ajay Mandlekar, Chaowei Xiao, Yuke Zhu, Linxi Fan, and Anima Anandkumar. 2023{\natexlab{a}}.
\newblock Voyager: An open-ended embodied agent with large language models.
\newblock \emph{arXiv preprint arXiv:2305.16291}.

\bibitem[{Wang et~al.(2023{\natexlab{b}})Wang, Li, Shao, Xu, Dai, Li, Chen, Wu, and Sui}]{wang2023math}
Peiyi Wang, Lei Li, Zhihong Shao, RX~Xu, Damai Dai, Yifei Li, Deli Chen, Y~Wu, and Zhifang Sui. 2023{\natexlab{b}}.
\newblock Math-shepherd: A label-free step-by-step verifier for llms in mathematical reasoning.
\newblock \emph{arXiv preprint arXiv:2312.08935}.

\bibitem[{Wang et~al.(2021)Wang, Chen, and Zhu}]{wang2021survey}
Xin Wang, Yudong Chen, and Wenwu Zhu. 2021.
\newblock A survey on curriculum learning.
\newblock \emph{IEEE transactions on pattern analysis and machine intelligence}, 44(9):4555--4576.

\bibitem[{Wang et~al.(2024)Wang, Li, Wu, Luo, Hou, Yu, and Shang}]{wang2024multi}
Zihan Wang, Yunxuan Li, Yuexin Wu, Liangchen Luo, Le~Hou, Hongkun Yu, and Jingbo Shang. 2024.
\newblock Multi-step problem solving through a verifier: An empirical analysis on model-induced process supervision.
\newblock \emph{arXiv preprint arXiv:2402.02658}.

\bibitem[{Wei et~al.(2022)Wei, Tay, Bommasani, Raffel, Zoph, Borgeaud, Yogatama, Bosma, Zhou, Metzler et~al.}]{wei2022emergent}
Jason Wei, Yi~Tay, Rishi Bommasani, Colin Raffel, Barret Zoph, Sebastian Borgeaud, Dani Yogatama, Maarten Bosma, Denny Zhou, Donald Metzler, et~al. 2022.
\newblock Emergent abilities of large language models.
\newblock \emph{arXiv preprint arXiv:2206.07682}.

\bibitem[{Xie et~al.(2023)Xie, Zhou, Cheng, Shi, Weng, Liu, Hua, Zhao, Liu, Liu et~al.}]{xie2023openagents}
Tianbao Xie, Fan Zhou, Zhoujun Cheng, Peng Shi, Luoxuan Weng, Yitao Liu, Toh~Jing Hua, Junning Zhao, Qian Liu, Che Liu, et~al. 2023.
\newblock Openagents: An open platform for language agents in the wild.
\newblock \emph{arXiv preprint arXiv:2310.10634}.

\bibitem[{Yang et~al.(2024)Yang, Prabhakar, Narasimhan, and Yao}]{yang2024intercode}
John Yang, Akshara Prabhakar, Karthik Narasimhan, and Shunyu Yao. 2024.
\newblock Intercode: Standardizing and benchmarking interactive coding with execution feedback.
\newblock \emph{Advances in Neural Information Processing Systems}, 36.

\bibitem[{Yao et~al.(2022{\natexlab{a}})Yao, Chen, Yang, and Narasimhan}]{yao2022webshop}
Shunyu Yao, Howard Chen, John Yang, and Karthik Narasimhan. 2022{\natexlab{a}}.
\newblock Webshop: Towards scalable real-world web interaction with grounded language agents.
\newblock \emph{Advances in Neural Information Processing Systems}, 35:20744--20757.

\bibitem[{Yao et~al.(2022{\natexlab{b}})Yao, Zhao, Yu, Du, Shafran, Narasimhan, and Cao}]{yao2022react}
Shunyu Yao, Jeffrey Zhao, Dian Yu, Nan Du, Izhak Shafran, Karthik Narasimhan, and Yuan Cao. 2022{\natexlab{b}}.
\newblock React: Synergizing reasoning and acting in language models.
\newblock \emph{arXiv preprint arXiv:2210.03629}.

\bibitem[{Yin et~al.(2023)Yin, Brahman, Ravichander, Chandu, Chang, Choi, and Lin}]{yin2023lumos}
Da~Yin, Faeze Brahman, Abhilasha Ravichander, Khyathi Chandu, Kai-Wei Chang, Yejin Choi, and Bill~Yuchen Lin. 2023.
\newblock Lumos: Learning agents with unified data, modular design, and open-source llms.
\newblock \emph{arXiv preprint arXiv:2311.05657}.

\bibitem[{Yu et~al.(2018)Yu, Zhang, Yang, Yasunaga, Wang, Li, Ma, Li, Yao, Roman et~al.}]{yu2018spider}
Tao Yu, Rui Zhang, Kai Yang, Michihiro Yasunaga, Dongxu Wang, Zifan Li, James Ma, Irene Li, Qingning Yao, Shanelle Roman, et~al. 2018.
\newblock Spider: A large-scale human-labeled dataset for complex and cross-domain semantic parsing and text-to-sql task.
\newblock \emph{arXiv preprint arXiv:1809.08887}.

\bibitem[{Yuan et~al.(2024)Yuan, Cui, Wang, Ding, Wang, Deng, Shan, Chen, Xie, Lin et~al.}]{yuan2024advancing}
Lifan Yuan, Ganqu Cui, Hanbin Wang, Ning Ding, Xingyao Wang, Jia Deng, Boji Shan, Huimin Chen, Ruobing Xie, Yankai Lin, et~al. 2024.
\newblock Advancing llm reasoning generalists with preference trees.
\newblock \emph{arXiv preprint arXiv:2404.02078}.

\bibitem[{Yuan et~al.(2023)Yuan, Yuan, Li, Dong, Tan, and Zhou}]{yuan2023scaling}
Zheng Yuan, Hongyi Yuan, Chengpeng Li, Guanting Dong, Chuanqi Tan, and Chang Zhou. 2023.
\newblock Scaling relationship on learning mathematical reasoning with large language models.
\newblock \emph{arXiv preprint arXiv:2308.01825}.

\bibitem[{Zelikman et~al.(2023)Zelikman, Lorch, Mackey, and Kalai}]{zelikman2023self}
Eric Zelikman, Eliana Lorch, Lester Mackey, and Adam~Tauman Kalai. 2023.
\newblock Self-taught optimizer (stop): Recursively self-improving code generation.
\newblock \emph{arXiv preprint arXiv:2310.02304}.

\bibitem[{Zeng et~al.(2023)Zeng, Liu, Lu, Wang, Liu, Dong, and Tang}]{zeng2023agenttuning}
Aohan Zeng, Mingdao Liu, Rui Lu, Bowen Wang, Xiao Liu, Yuxiao Dong, and Jie Tang. 2023.
\newblock Agenttuning: Enabling generalized agent abilities for llms.
\newblock \emph{arXiv preprint arXiv:2310.12823}.

\bibitem[{Zhu et~al.(2024)Zhu, Qiao, Ou, Deng, Zhang, Lyu, Shen, Liang, Gu, and Chen}]{zhu2024knowagent}
Yuqi Zhu, Shuofei Qiao, Yixin Ou, Shumin Deng, Ningyu Zhang, Shiwei Lyu, Yue Shen, Lei Liang, Jinjie Gu, and Huajun Chen. 2024.
\newblock Knowagent: Knowledge-augmented planning for llm-based agents.
\newblock \emph{arXiv preprint arXiv:2403.03101}.

\end{thebibliography}

\clearpage

\appendix

\section{Dataset Details}
\label{appendix:datasets}
\paragraph{WebShop} 
WebShop~\citep{yao2022webshop} is a network-based simulation environment for e-commerce experiences, features a website with 1.8 million actual products, each with distinct labels and attributes. In this environment, the agent is allowed to interact with the system through "search[QUERY]" or "click[ELEMENT]" actions to purchase products matching the instructions. Once the agent clicks the "buy" option, the environment provides a final reward, which is calculated based on the matching heuristics of the product's attributes and price.

\paragraph{InterCodeSQL}
InterCodeSQL is an interactive database environment within InterCode benchmark~\citep{yang2024intercode}, where the agent interacts with the environment to retrieve necessary table information and complete the corresponding SQL queries. The database is constructed from the Spider~\citep{yu2018spider} dataset, a large-scale cross-domain dataset originally designed for evaluating SQL query generation from natural language questions. We have modified InterCodeSQL to fit for our evaluation framework. When the agent perform the "submit" action, the environment provides a final reward. The reward is calculated using the Intersection over Union~(\textit{IoU}) metric to quantify the correctness of the submitted execution output generated by the against the gold output, with both outputs being lists of records.

\paragraph{ALFWorld}
ALFWorld~\citep{shridhar2020alfworld} are household tasks that require agents to explore rooms and use commonsense reasoning to perform tasks, such as "put a pencil on the desk". The environment provides the outcome on whether the agent successfully completes the task within given steps. The original ALFWorld dataset comprises both seen and unseen evaluation sets. The seen set is designed to assess in-distribution generalization, whereas the unseen set with new task instances measures out-of-distribution generalization of the agents.

\section{Details of the Scoring Function}
\label{appendix:scoring}
In the WebShop environment, \citet{yao2022webshop} provides the scoring formula to calculate the score of any product (the distance from the target product) as follows:
\begin{equation}
\scalebox{1}{$
    f = f_{\mathrm{type}} \cdot \frac{|\mathcal U_{\mathrm{att}} \cap \mathcal Y_{\mathrm{att}}| + |\mathcal U_{\mathrm{opt}} \cap \mathcal Y_{\mathrm{opt}}| + \textbf{1}[\mathrm y_{\mathrm{price}} \leq \mathrm u_{\mathrm{price}}]}{|\mathcal U_{\mathrm{att}}| + |\mathcal U_{\mathrm{opt}}| + 1}, 
$}
\end{equation}
where $f_{\mathrm{type}} = \mathrm{TextMatch} (\overline{y}, \overline{y}^*)$. Following ~\citet{ma2024agentboard}, we expand the product scoring rules to derive the score for each action. Typically, completing a web shopping task involves three continuous states: search, product selection, and finalizing the product style before placing an order. Each action leads to deterministic state change in the environment. Therefore, to calculate the step reward, we measure the distance between the result state and the target state. We primarily calculate scores for three pages~(states): search result page, product description page, and order confirmation page. On the search result page, we calculate the score of each product on the page and take the highest score for this page. On the product description page, we compute the highest score for the product under various options as the page score. On the order confirmation page, the score of the finally selected product is considered as the score for that page.

\section{Training Efficiency Analysis}
\label{appendix:training efficiency}
Here, we compare the time consumption of different methods on WebShop in Figure~\ref{fig:overall}. Since our method can achieve state-of-the-art performance after three rounds of iteration, we use the time for three rounds of iteration as the measure of training time. The time consumption results are as follows: SFT requires 1 hour, ETO requires 2.5 hours, and IPR requires 5.3 hours. Furthermore, although the Monte Carlo method necessitates sampling to obtain the process information of step rewards, with the support of vllm~\citep{kwon2023efficient}, we have indeed been able to construct the step rewards in an efficient and parallel manner. Without increasing inference time costs, IPR achieves nearly a 6\% performance improvement at the expense of a training duration less than three times that of ETO. We believe that this time cost is acceptable.

% \begin{table}[]
%     \centering
%     \begin{tabular}{l | c}
%     \toprule
%     \textbf{Method} & \textbf{Training Time} \\
%     \midrule
%     SFT  &  1h \\
%     ETO  &  2.5h \\
%     IPR & 5.3h \\
%     \bottomrule
%     \end{tabular}
%     \caption{Training time consumption.}
%     \label{tab:training time}
% \end{table}

\section{Case Study}
\label{appendix:case study}
Here, we provide a detailed comparison of the trajectories of \ourmethod{} and ETO on the same tasks within WebShop and ALFWorld, which illustrates how \ourmethod{} provides fine-grained guidance to correct erroneous actions and improves the efficacy of agent actions. The cases are shown in Figure~\ref{fig:webshop_case} and ~\ref{fig:alfworld_case}.

In the WebShop scenario, ETO learned the process of selecting products and choosing their attributes. However, when selecting a product, it overlooked the requirement for product type and directly purchased the first item on the search page, a \texttt{dress shirt}, ignoring the request for a \texttt{tank top}. In contrast, our \ourmethod{}, benefiting from fine-grained process supervision, corrected this error and chose the correct \texttt{tank top} product.

In the ALFWorld scenario, ETO engaged in blind exploration to find the desired item \texttt{potato}, which introduced a lot of irrelevant information and unnecessary context. In contrast, our \ourmethod{} improved the efficiency of each action, directly navigating to the \texttt{sinkbasin} to locate the \texttt{potato}. Moreover, ETO made a mistake in the details by cooling the \texttt{potato} down without picking it up, whereas \ourmethod{} rectified this action.

\onecolumn
\begin{tcolorbox}[breakable, title=Case Study of WebShop, enhanced jigsaw]
\columnseprule=0.5pt
\begin{multicols}{2}
\begin{center}\textcolor{red}{\textbf{\Large ETO}}\end{center}
\textcolor{case_purple}{\textbf{Task Instruction:}} I want to find a small purple bike tank top for men that has a classic fit, and price lower than 50.00 dollars\\
\textcolor{case_blue}{\textbf{Thought:}} I need to search for a small purple bike tank top for men that has a classic fit and is priced under \$50.\\
\textcolor{case_blue}{\textbf{Action:}} search[small purple bike tank top men classic fit]\\
\textcolor{case_purple}{\textbf{Observation:}} Page 1 (Total results: 50) [SEP] Next > [SEP] B09PBY1KJG [SEP] Men's Vertical Striped Button Down Dress Shirt Long Sleeve Casual Comfort Regular Fit Business Oxford Shirts [SEP] \$1.97 to \$22.97 ... [SEP] B08W4FVVXC [SEP] Just Send It Motocross Cool Motorcycle Dirt Bike Biker Gift Tank Top [SEP] \$19.9 ... \\
\sethlcolor{red!50}\hl{\textit{\# Overlooking important requirements}}\\
\textcolor{case_blue}{\textbf{Thought:}} The search results are displayed. \sethlcolor{red!50}\hl{The first item, B09PBY1KJG, is a men's vertical striped button-down dress shirt that fits the criteria.} I should click on it to see more details and confirm if it's the right size and price.\\
\textcolor{case_blue}{\textbf{Action:}} click[B09PBY1KJG]\\
\textcolor{case_purple}{\textbf{Observation:}} \textit{<product details>}\\
\textcolor{case_blue}{\textbf{Thought:}} The shirt is available in purple, which is the color I want. I should select this color option.\\
\textcolor{case_blue}{\textbf{Action:}} click[purple]\\
\textcolor{case_purple}{\textbf{Observation:}} \textit{<product details>}\\
\textcolor{case_blue}{\textbf{Thought:}} Now I need to ensure the shirt is available in small size, as that's the size I'm looking for.\\
\textcolor{case_blue}{\textbf{Action:}} click[small]\\
\textcolor{case_purple}{\textbf{Observation:}} \textit{<product details>}\\
\textcolor{case_blue}{\textbf{Thought:}} \sethlcolor{red!50}\hl{The shirt is purple, small-sized, and within my budget. It meets all the criteria I was looking for.} I should go ahead and purchase this item.\\
\textcolor{case_blue}{\textbf{Action:}} click[Buy Now]\\
\textcolor{case_purple}{\textbf{Reward:}} 0.0\\
\columnbreak

\begin{center}\textcolor{my_green}{\textbf{\Large IPR}}\end{center}
\textcolor{case_purple}{\textbf{Task Instruction:}} I want to find a small purple bike tank top for men that has a classic fit, and price lower than 50.00 dollars.\\
\textcolor{case_blue}{\textbf{Thought:}} I think I should use the search bar to look for the product I need.\\
\textcolor{case_blue}{\textbf{Action:}} search[small purple bike tank top men classic fit]\\
\textcolor{case_purple}{\textbf{Observation:}} Page 1 (Total results: 50) [SEP] Next > [SEP] B09PBY1KJG [SEP] Men's Vertical Striped Button Down Dress Shirt Long Sleeve Casual Comfort Regular Fit Business Oxford Shirts [SEP] \$1.97 to \$22.97 ... [SEP] B08W4FVVXC [SEP] Just Send It Motocross Cool Motorcycle Dirt Bike Biker Gift Tank Top [SEP] \$19.9 ...\\
\textcolor{case_blue}{\textbf{Thought:}} \sethlcolor{green!50}\hl{I think I should click on the product that matches my criteria, which is B08W4FVVXC.}\\
\textcolor{case_blue}{\textbf{Action:}} \sethlcolor{green!50}\hl{click[B08W4FVVXC]}\\
\textcolor{case_purple}{\textbf{Observation:}} \textit{<product details>}\\
\textcolor{case_blue}{\textbf{Thought:}} I think I should click on the 'purple' color option to select the color I need.\\
\textcolor{case_blue}{\textbf{Action:}} click[purple]\\
\textcolor{case_purple}{\textbf{Observation:}} \textit{<product details>}\\
\textcolor{case_blue}{\textbf{Thought:}} I think I should click on the 'small' size option to select the size I need.\\
\textcolor{case_blue}{\textbf{Action:}} click[small]\\
\textcolor{case_purple}{\textbf{Observation:}} \textit{<product details>}\\
\textcolor{case_blue}{\textbf{Thought:}} I think I should click on the 'Buy Now' button to purchase the product.\\
\textcolor{case_blue}{\textbf{Action:}} click[Buy Now]\\
\textcolor{case_purple}{\textbf{Reward:}} 1.0\\
\end{multicols}
\end{tcolorbox}
\begin{figure}[!h]
    \centering
    \vspace{-8pt}
    \caption{
    Case study for WebShop.
    }
    \label{fig:webshop_case}
\end{figure}

\begin{tcolorbox}[breakable, title=Case Study of ALFWorld, enhanced jigsaw]
\columnseprule=0.5pt
\begin{multicols}{2}
\begin{center}\textcolor{red}{\textbf{\Large ETO}}\end{center}
\textcolor{case_purple}{\textbf{Task Instruction:}} You are in the middle of a room. Looking quickly around you, you see a cabinet 6, a cabinet 5, a cabinet 4, a cabinet 3, a cabinet 2, a cabinet 1, a coffeemachine 1, a countertop 3, a countertop 2, a countertop 1, a drawer 3, a drawer 2, a drawer 1, a fridge 1, a garbagecan 1, a microwave 1, a shelf 3, a shelf 2, a shelf 1, a sinkbasin 1, a stoveburner 4, a stoveburner 3, a stoveburner 2, a stoveburner 1, and a toaster 1. Your task is to: cool some potato and put it in microwave.\\
\sethlcolor{red!50}\hl{\textit{\# Actions of blind exploration}}\\
\textcolor{case_blue}{\textbf{Thought:}} I need to cool a potato before putting it in the microwave. The fridge is the best place to cool food items quickly.\\
\textcolor{case_blue}{\textbf{Action:}} go to fridge 1\\
\textcolor{case_purple}{\textbf{Observation:}} The fridge 1 is closed.\\
\textcolor{case_blue}{\textbf{Thought:}} The fridge is closed. I need to open it to access the cooling space inside.\\
\textcolor{case_blue}{\textbf{Action:}} open fridge 1\\
\textcolor{case_purple}{\textbf{Observation:}} You open the fridge 1. The fridge 1 is open. In it, you see a bowl 1, a cup 1, a lettuce 1, a plate 2, and a tomato 1.\\
\textcolor{case_blue}{\textbf{Thought:}} I've opened the fridge, but I realize I don't need to put the potato inside since I can cool it while holding it. I should close the fridge to conserve energy.\\
\textcolor{case_blue}{\textbf{Action:}} close fridge 1\\
\textcolor{case_purple}{\textbf{Observation:}} You close the fridge 1.\\
\textcolor{case_blue}{\textbf{Thought:}} I need to cool the potato, but I can't use the fridge since I closed it. I should find another way to cool it down quickly.\\
\textcolor{case_blue}{\textbf{Action:}} go to sinkbasin 1\\
\textcolor{case_purple}{\textbf{Observation:}} On the sinkbasin 1, you see a bowl 3, a butterknife 2, a egg 2, a fork 2, a knife 1, and a potato 2.\\
\sethlcolor{red!50}\hl{\textit{\# Making a mistake in the details by cooling down without picking up the potato}}\\
\textcolor{case_blue}{\textbf{Thought:}} I'm at the sink basin, which is a common place to cool food items by placing them in cold water. I can use this method to cool the potato.\\
\textcolor{case_blue}{\textbf{Action:}} cool potato 2 with sinkbasin 1\\
\textcolor{case_purple}{\textbf{Observation:}} Nothing happens.\\
\textcolor{case_blue}{\textbf{Thought:}} Now that the potato is cooled, I need to put it in the microwave. I should go to the microwave to proceed with the task.\\

\columnbreak

\begin{center}\textcolor{my_green}{\textbf{\Large IPR}}\end{center}
\textcolor{case_purple}{\textbf{Task Instruction:}} You are in the middle of a room. Looking quickly around you, you see a cabinet 6, a cabinet 5, a cabinet 4, a cabinet 3, a cabinet 2, a cabinet 1, a coffeemachine 1, a countertop 3, a countertop 2, a countertop 1, a drawer 3, a drawer 2, a drawer 1, a fridge 1, a garbagecan 1, a microwave 1, a shelf 3, a shelf 2, a shelf 1, a sinkbasin 1, a stoveburner 4, a stoveburner 3, a stoveburner 2, a stoveburner 1, and a toaster 1. Your task is to: cool some potato and put it in microwave.\\
\textcolor{case_blue}{\textbf{Thought:}} I need to cool a potato and then put it in the microwave. First, I should locate the potato.\\
\textcolor{case_blue}{\textbf{Action:}} go to sinkbasin 1\\
\textcolor{case_purple}{\textbf{Observation:}} On the sinkbasin 1, you see a bowl 3, a butterknife 2, a egg 2, a fork 2, a knife 1, and a potato 2.\\
\textcolor{case_blue}{\textbf{Thought:}} There's a potato here on the sink basin. I should pick it up to proceed with the task.\\
\textcolor{case_blue}{\textbf{Action:}} take potato 2 from sinkbasin 1\\
\textcolor{case_purple}{\textbf{Observation:}} You pick up the potato 2 from the sinkbasin 1.\\
\textcolor{case_blue}{\textbf{Thought:}} To cool the potato, I need to use the fridge. I should go to the fridge now.\\
\textcolor{case_blue}{\textbf{Action:}} go to fridge 1\\
\textcolor{case_purple}{\textbf{Observation:}} The fridge 1 is closed.\\
\textcolor{case_blue}{\textbf{Thought:}} The fridge is closed. I need to open it to cool the potato inside.\\
\textcolor{case_blue}{\textbf{Action:}} open fridge 1\\
\textcolor{case_purple}{\textbf{Observation:}} You open the fridge 1. The fridge 1 is open. In it, you see a bowl 1, a cup 1, a lettuce 1, a plate 2, and a tomato 1.\\
\textcolor{case_blue}{\textbf{Thought:}} The fridge is open now. I can use it to cool the potato by placing the potato inside.\\
\textcolor{case_blue}{\textbf{Action:}} cool potato 2 with fridge 1\\
\textcolor{case_blue}{\textbf{Thought:}} The potato is cooled. The next step is to put it in the microwave, so I need to go to the microwave.\\
\textcolor{case_blue}{\textbf{Action:}} go to microwave 1\\
\textcolor{case_purple}{\textbf{Observation:}} The microwave 1 is closed.\\
\textcolor{case_blue}{\textbf{Thought:}} The microwave is closed. I need to open it to place the potato inside.\\
\textcolor{case_blue}{\textbf{Action:}} open microwave 1\\
\textcolor{case_purple}{\textbf{Observation:}} You open the microwave 1. The microwave 1 is open. In it, you see a egg 1.\\
\end{multicols}

\begin{multicols}{2}
\textcolor{case_blue}{\textbf{Action:}} go to microwave 1\\
\textcolor{case_purple}{\textbf{Observation:}} The microwave 1 is closed.\\
\textcolor{case_blue}{\textbf{Thought:}} The microwave is closed. I need to open it to place the potato inside.\\
\textcolor{case_blue}{\textbf{Action:}} open microwave 1\\
\textcolor{case_purple}{\textbf{Observation:}} You open the microwave 1. The microwave 1 is open. In it, you see a egg 1.\\
\textcolor{case_blue}{\textbf{Thought:}} I'll proceed with the given instructions and place the potato in the microwave.\\
\textcolor{case_blue}{\textbf{Action:}} put potato 2 in/on microwave 1\\
\textcolor{case_purple}{\textbf{Observation:}} Nothing happens\\
\textcolor{case_blue}{\textbf{Thought:}} With the potato in the microwave, the next step is to close it to complete the task.\\
\textcolor{case_blue}{\textbf{Action:}} close microwave 1\\
\textcolor{case_purple}{\textbf{Observation:}} You close the microwave 1.\\
\textcolor{case_blue}{\textbf{Thought:}} The task is complete with the potato placed in the microwave. I should ensure the microwave is closed before finishing.\\
\textcolor{case_blue}{\textbf{Action:}} put potato 2 in/on microwave 1\\
\textcolor{case_purple}{\textbf{Observation:}} You cool the potato 2 using the fridge 1.\\
....\\
\textcolor{case_purple}{\textbf{Reward:}} 0.0\\
\columnbreak

\textcolor{case_blue}{\textbf{Thought:}} The microwave is open, I'll proceed with placing the potato in as instructed.\\
\textcolor{case_blue}{\textbf{Action:}} put potato 2 in/on microwave 1\\
\textcolor{case_purple}{\textbf{Observation:}} You put the potato 2 in/on the microwave 1.\\
\textcolor{case_purple}{\textbf{Reward:}} 1.0\\

\end{multicols}

\end{tcolorbox}
\begin{figure}[!h]
    \centering
    \vspace{-8pt}
    \caption{
    Case study for ALFWorld.
    }
    \label{fig:alfworld_case}
\end{figure}

\twocolumn

% \section{Training Data Example}
% \label{appendix:training}
% We show the training data example in Figure~\ref{fig:training_data}.

\section{Expert Trajectories Collection}
\label{appendix:collection}
We primarily us the expert trajectories collected by~\citet{song2024trial} in ReAct pattern. For InterCodeSQL tasks not covered by these trajectories, we conducted our annotations.
\begin{itemize}
    \item \textbf{WebShop}~\citep{yao2022webshop}. In addition to manually annotated trajectories provided by the WebShop, GPT-4 is employed to annotate additional trajectories. The trajectories with final rewards exceeding 0.7 are reserved.
    \item  \textbf{InterCodeSQL}~\citep{yang2024intercode}. We annotate expert trajectories using GPT-4 and retain trajectories with a reward of 1.0.
    \item \textbf{ALFWorld}~\citep{shridhar2020alfworld}. The dataset provides human-annotated trajectories.
\end{itemize}
As the original trajectories lack the thoughts for each action step, we have employed GPT-4 to generate the corresponding information.

\section{Prompt for Evaluation}
\label{appendix:prompt}
We show the instruction prompts for WebShop, InterCodeSQL, ALFWorld in Figure~\ref{fig:webshop_prompt}, \ref{fig:intercodesql_prompt}, \ref{fig:alfworld_prompt}, respectively.

\onecolumn
\tcbset{width=\textwidth}

% \begin{tcolorbox}[breakable,title=Training Data Example for WebShop ]
% \textcolor{case_purple}{\textbf{Task Instruction:} WebShop [SEP] Instruction: [SEP] i need gluten free vegetarian smoked peppered bacon - 4 ounce (pack of 2), and price lower than 40.00 dollars [SEP] Search} \\
% \textcolor{case_blue}{\textbf{Thought:} I need to find gluten-free vegetarian smoked peppered bacon that comes in a 4-ounce pack of 2 and is under \$40. I'll start by searching for smoked peppered bacon in the specified size.} \\
% \textcolor{case_blue}{\textbf{Action:} search[smoked peppered bacon - 4 ounce (pack of 2)]}\\
% \textcolor{case_purple}{\textbf{Observation:} Page 1 (Total results: 50) [SEP] Next > [SEP] B06Y96MXJV [SEP] Smoked Bacon Sea Salt 3-Pack: Smoked Bacon Chipotle, Smoked Bacon and Onion, Smoked Peppered Bacon - All-Natural Sea Salts Slowly Smoked Over Alderwood - No Gluten, No MSG, Non-GMO (12 total oz.) [SEP] \$29.99 ...} \\
% \textcolor{case_blue}{\textbf{Thought:} he search results are in. B06Y96MXJV looks like it could be the product I'm looking for, as it's a smoked bacon sea salt pack and is priced under \$40. I should click on it to see more details and confirm if it's gluten-free and vegetarian.}
% \end{tcolorbox}
% \begin{figure*}[ht]
%     \centering
%     \vspace{-8pt}
%     \caption{
%     Training Data Example for WebShop.
%     }
%     \label{fig:training_data}
% \end{figure*}

\begin{tcolorbox}[breakable,title=Instruction Prompt for WebShop ]
You are doing a web shopping task.
I will give you instructions about what to do.
You have to follow the instructions. Every round I will give you an observation and a list of available actions, you have to respond to an action based on the state and instruction.
You can use search action if search is available.
You can click one of the buttons in clickables.
An action should be one of the following structure: search[keywords] or click[value]\\

If the action is not valid, perform nothing.
Keywords in search are up to you, but the value in click must be a value in the list of available actions.
Remember that your keywords in search should be carefully designed.\\

Your response should use the following format:\\
Thought: I think ...\\
Action: click[something]
\end{tcolorbox}
\begin{figure*}[ht]
    \centering
    \vspace{-8pt}
    \caption{
    Instruction prompt for WebShop.
    }
    \label{fig:webshop_prompt}
\end{figure*}

\begin{tcolorbox}[breakable, title=Instruction Prompt for InterCodeSQL, enhanced jigsaw]
You are a helpful assistant assigned with the task of problem-solving. To achieve this, you will interact with a MySQL Database system using SQL queries to answer a question.

At each turn, you should first provide your step-by-step thinking for solving the task. Your thought process should start with "Thought: ", for example: Thought: I should write a SQL query that gets the average GNP and total population from nations whose government is US territory. \\

After that, you have two options:

1) Interact with a mysql programming environment and receive the corresponding output. Your code should start with "Action: " , for example: Action:  SELECT AVG(GNP), SUM(population) FROM nations WHERE government = `US Territory'

2) Directly submit the result, for example: Action: submit. \\

You should use this format: \\ Thought: your thought\\ Action: <the mysql command>. \\

You will receive the corresponding output for your sql command.
Your output should contain only one "Action" part.
The "Action" part should be executed with a mysql interpreter or propose an answer. Any natural language in it should be commented out.
The SQL query and submit parts can not appear in your output simultaneously.
\end{tcolorbox}
\begin{figure*}[ht]
    \centering
    \vspace{-8pt}
    \caption{
    Instruction prompt for InterCodeSQL.
    }
    \label{fig:intercodesql_prompt}
\end{figure*}

\begin{tcolorbox}[breakable, title=Instruction Prompt for ALFWorld, enhanced jigsaw]
Interact with a household to solve a task. Imagine you are an intelligent agent in a household environment and your target is to perform actions to complete the task goal. At the beginning of your interactions, you will be given a detailed description of the current environment and your goal to accomplish. \\
For each of your turn, you will be given the observation of the last turn. You should first think about the current condition and plan for your future actions, and then output your action in this turn. Your output must strictly follow this format:"Thought: your thoughts. Action: your next action".\\

The available actions are:\\
1. go to {recep}\\
2. task {obj} from {recep}\\
3. put {obj} in/on {recep}\\
4. open {recep}\\
5. close {recep}\\
6. toggle {obj} {recep}\\
7. clean {obj} with {recep}\\
8. heat {obj} with {recep}\\
9. cool {obj} with {recep}\\
where {obj} and {recep} correspond to objects and receptacles.\\
After each turn, the environment will give you immediate feedback based on which you plan your next few steps. if the environment outputs "Nothing happened", that means the previous action is invalid and you should try more options.\\

Your response should use the following format:\\
Thought: <your thoughts>\\
Action: <your next action>
\end{tcolorbox}
\begin{figure*}[ht]
    \centering
    \vspace{-8pt}
    \caption{
    Instruction prompt for ALFWorld.
    }
    \label{fig:alfworld_prompt}
\end{figure*}

\twocolumn

\end{document}